\documentclass[11pt]{article}

\PassOptionsToPackage{table}{xcolor}

\usepackage[preprint]{acl}

\usepackage{times}
\usepackage{latexsym}

\usepackage[T1]{fontenc}

\usepackage[utf8]{inputenc}

\usepackage{microtype}

\usepackage{amsmath}
\usepackage{graphicx}
\usepackage{array}
\usepackage{tabularx}
\usepackage{ragged2e}
\usepackage{booktabs}
\usepackage{enumitem}
\usepackage{inconsolata}

\newcolumntype{Y}{>{\RaggedRight\arraybackslash}X}

\definecolor{softblue}{RGB}{240, 248, 255}
\definecolor{softpeach}{RGB}{253, 242, 235}

\title{Omanic: Towards Step-wise Evaluation of Multi-hop Reasoning \\ in Large Language Models}

\author{
  \textbf{Xiaojie Gu}\textsuperscript{1},
  \textbf{Sherry T. Tong}\textsuperscript{1},
  \textbf{Aosong Feng}\textsuperscript{2},
  \textbf{Sophia Simeng Han}\textsuperscript{3},\\
  \textbf{Jinghui Lu}\textsuperscript{4,5},
  \textbf{Yingjian Chen}\textsuperscript{1},
  \textbf{Yusuke Iwasawa}\textsuperscript{1},
  \textbf{Yutaka Matsuo}\textsuperscript{1},\\
  \textbf{Chanjun Park}\textsuperscript{6},
  \textbf{Rex Ying}\textsuperscript{2},
  \textbf{Irene Li}\textsuperscript{1,5\ensuremath{\dagger}} \\
  \textsuperscript{1}The University of Tokyo,
  \textsuperscript{2}Yale University,
  \textsuperscript{3}Stanford University,\\
  \textsuperscript{4}Xiaomi EV, 
\textsuperscript{5}Smartor LLC, 
  \textsuperscript{6}Soongsil University\\
  $^{\dagger}$\small{\textbf{Corresponding author:}
\texttt{irene.li@weblab.t.u-tokyo.ac.jp}}
}

\begin{document}
\maketitle
\begin{abstract}

Evaluating the reasoning abilities of large language models (LLMs) solely from final answers can obscure failures in intermediate steps, especially in multi-hop QA benchmarks without step-level annotations.
To address this gap, we introduce Omanic, an open-domain 4-hop QA benchmark designed not only to measure final-answer accuracy but also to diagnose where reasoning breaks down.
Omanic contains 10,296 machine-generated training examples (OmanicSynth) and 967 expert-reviewed human-annotated evaluation examples (OmanicBench), with each evaluation question decomposed into single-hop sub-questions, intermediate answers, and structured graph topologies.
Experiments with proprietary and open-source LLMs show that Omanic is challenging, while step-wise analysis reveals a later-hop bottleneck, factual knowledge floor, and error propagation along reasoning chains.
Fine-tuning on OmanicSynth transfers to six reasoning and mathematics benchmarks, yielding a 7.41-point average gain and validating its effectiveness as supervision for reasoning-capability transfer.
We release the data at \url{https://huggingface.co/datasets/li-lab/Omanic}.\footnote{Code at \url{https://github.com/XiaojieGu/Omanic}.}


\end{abstract}

\section{Introduction}


As LLMs mature~\cite{gemini3-pro,gpt5-1,qwen3maxthinking}, the research frontier has shifted from single-task proficiency to complex reasoning, with Chain-of-Thought (CoT) prompting~\cite{cot} becoming central for eliciting intermediate steps. Yet growing evidence shows that LLMs often rely on \textbf{\textit{reasoning shortcuts}}~\cite{illusion_thinking,gao2026med,Reasoning_Trace}, reaching correct answers through heuristic pattern matching rather than rigorous deduction, while high end-to-end accuracy can hide failures in intermediate steps~\cite{verify_cot}. 
This concern is acute in multi-hop reasoning: benchmarks such as MuSiQue~\cite{musique} advance cross-document evidence synthesis, but typically evaluate only final answers and lack step-level annotations for diagnosing \textit{where} and \textit{why} models fail. Without such ground truth, it remains difficult to distinguish genuine compositional reasoning from shortcut exploitation~\cite{synthworlds}.

To bridge this gap, we introduce \textbf{OmanicBench} (\textbf{O}pen-domain \textbf{M}ulti-hop questions with \textbf{AN}notated reason\textbf{I}ng \textbf{C}hain), a 4-hop open-domain QA benchmark centered on step-wise diagnosis rather than final-answer accuracy alone.
OmanicBench consists of 967 multi-hop questions, each manually annotated and expert-reviewed (over 300 hours of annotation effort), providing high-quality ground truth for analyzing reasoning behavior. Crucially, every question is decomposed into four cross-domain single-hop sub-questions with intermediate answers, drawing on rich factual knowledge and connected through mathematical reasoning. These annotations make it possible to evaluate each hop, inspect how answers propagate, and identify where a model's reasoning chain breaks.
Besides, we also release OmanicSynth, a machine-generated training set containing 10,296 instances for supervised training and transfer experiments.

\renewcommand{\arraystretch}{1.25}

\begin{table*}[t]
\centering
\scriptsize
\setlength{\tabcolsep}{6pt}

\begin{tabularx}{\textwidth}{
  >{\centering\arraybackslash}p{0.18\textwidth}
  p{0.35\textwidth}
  Y
}

\hline
\textbf{Reasoning Graph} & \textbf{Multi-hop Question} & \textbf{Single-hop Composition} \\

\hline

\vspace{0pt}\includegraphics[width=0.16\textwidth,keepaspectratio]{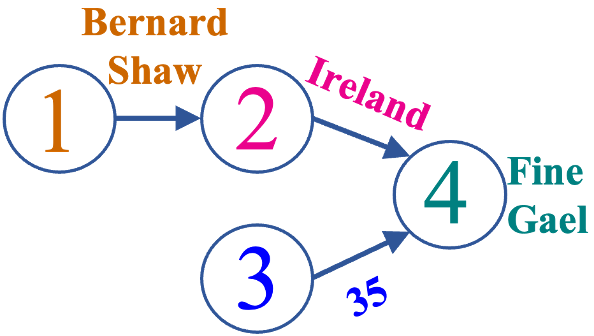}

&
\vspace{0pt}
In the country of citizenship of the author of Candida, which political party was founded the same number of years before 1968 as the number of distinct 3-member committees that can be formed from a group of 7 candidates?
\textcolor{teal}{Fine Gael}

\noindent\textit{[A: Fine Gael B: Labour Party C: Fianna Fáil D: Sinn Féin]}
&

\begin{enumerate}[leftmargin=*,nosep,topsep=0pt,partopsep=0pt]

    \item Who is the author of Candida?
    \textcolor{orange!80!black}{Bernard Shaw}
    \item What is the country of citizenship of \textcolor{orange!80!black}{Bernard Shaw}?
    \textcolor{magenta}{Ireland}
    \item How many distinct 3-member committees can be formed from a group of 7 candidates?
    \textcolor{blue}{35}
    \item In \textcolor{magenta}{Ireland}, which political party was founded \textcolor{blue}{35} years before 1968?
    \textcolor{teal}{Fine Gael}
\end{enumerate}
\\
\hline

\end{tabularx}
\caption{An example illustrating the multi-hop reasoning graph and its step-by-step question decomposition.}
\label{example_main}
\end{table*}

Leveraging the step-wise annotations in Omanic, we analyze state-of-the-art LLMs and reveal two phenomena obscured by final-answer evaluation. CoT gains exhibit a \textit{knowledge floor}, diminishing as required atomic facts become missing, while errors propagate along reasoning chains, with later hops consistently showing higher error rates. 
We further conduct an entropy-reduction analysis to probe these behaviors.
Together, these findings show that OmanicBench functions as a diagnostic testbed for separating factual retrieval, compositional reasoning, and propagation failures.

In summary, we make three main contributions. First, we propose Omanic, an open-domain 4-hop QA resource with 10,296 training and 967 expert-reviewed testing instances, where state-of-the-art LLMs only achieve 73.11\% accuracy. Second, fine-tuning on OmanicSynth yields a 7.41-point average gain across six external benchmarks, showing strong transferability and data quality. Third, using single-hop decomposition and intermediate answers, we empirically analyze the \textit{knowledge floor} effect and \textit{error propagation}, showing how OmanicBench localizes failures that final-answer evaluation would obscure.




\section{Omanic Construction Pipeline}

We describe the main dataset construction steps in this section. Figure~\ref{pipeline} in Appendix~\ref{Guidance} provides an overview of the full pipeline.

\paragraph{Triplets Retrieval}

To construct the Omanic dataset, we begin with the answers to the original 2-hop questions in MuSiQue \cite{musique}, which are themselves composed of two single-hop questions. These answers serve as anchor subjects for retrieving corresponding $(subject, relation, object)$ triplets from Wikidata5M \cite{Wikidata5m}, a large-scale knowledge graph derived from Wikipedia.
These retrieved triplets function as the foundational building blocks for our 4-hop expansion.

\paragraph{Constrained Synthesis}
To construct 4-hop queries, we merge original MuSiQue components with new single-hop questions synthesized from retrieved triplets, utilizing Claude-Sonnet-4.5.
This process is governed by domain constraints and reasoning-graph topology. Each synthesized single-hop question is assigned to one of eight predefined domains (e.g., \textit{History and Literature}, \textit{Art and Architecture}), and the source Wikidata triplet is contextually refined to improve fluency and coherence.
Each 4-hop instance is also required to contain at least one mathematically grounded hop. Numerical, temporal, or countable attributes are rewritten into sub-questions that require explicit quantitative reasoning, such as comparison, aggregation, counting, arithmetic composition, or temporal calculation. This mathematical hop is embedded into the chain rather than appended independently, so its inputs depend on earlier hops and its output can support later ones.
For each query, we randomly choose one of three reasoning graph topologies~\cite{musique} (Figure~\ref{reasoning-graphs}), which determines both question synthesis and final assembly. For example, under the Bridge pattern (Table~\ref{example_main}), the second hop depends on the first, and the fourth depends on both the second and third, preventing shortcut solutions that bypass intermediate reasoning. Finally, each single-hop question is paired with three distractors, and the answer and options of the last hop are used as the ground truth and candidate set for the full 4-hop query.

\paragraph{Automated Filtering}

To maintain a great difficulty ceiling, we filter the synthesized dataset using an ensemble of four models\footnote{Including Llama-3.1-8B-Instruct,  Qwen3-8B, Mistral-7B-Instruct-v0.3, and Gemma-3-4b-it.}. Any question answered correctly by two or more models is deemed too simple and discarded. 
After pruning, 3,415 instances were removed, yielding a final training set of 10,296 examples.

\paragraph{Expert Review}

To ensure the high quality of the OmanicBench, 1,172 candidate instances undergo rigorous human audit conducted by 10 trained undergraduates and postgraduates over approximately 300 person-hours via the \texttt{Label Studio} platform\footnote{\url{https://labelstud.io/}}.
For each instance, annotators first verify the factual correctness of every sub-question and its answer by consulting the Wikipedia articles linked to the underlying triplets; for questions involving mathematical reasoning, annotators are additionally required to provide detailed step-by-step computations.
They then examine the logical coherence of the full 4-hop reasoning chain, checking whether it conforms to the designated graph topology, and then assign quality scores across multiple dimensions (factual accuracy, distractor plausibility, fluency, and reasoning integrity) following a standardized rubric (detailed in Appendix~\ref{Guidance}). They also verify the correctness of the required numerical operations in mathematically grounded hops, ensuring that intermediate calculations and final derived values are arithmetically consistent.
Where deficiencies are identified, annotators directly correct the instance or supplement supporting references and derivations.
Instances failing to meet predefined quality thresholds were excluded, yielding a final set of 967 high-quality instances as the eval set.
An example instance is shown in Table~\ref{example_main}, and a comparison with previous benchmarks is provided in Table~\ref{tab:benchmark-comparison}.


\section{Experiments and Analysis}

\begin{table}[t]
    \centering
    \resizebox{\columnwidth}{!}{
    \begin{tabular}{lccc}
    \toprule
          & \textbf{MCQ} & \textbf{Exact Match} & \textbf{F1-Score}   \\
        \midrule
         \textit{\textbf{Proprietary LLMs}} \\
 GPT-5.4                              & 49.22 & 22.85 & 32.22 \\
  \rowcolor{softpeach}
 GPT-5.4$_\texttt{CoT}$                 & 70.84 & 27.09 & 43.58 \\
 Claude-Sonnet-4.6                    & 55.43 & 20.73 & 37.15 \\
  \rowcolor{softpeach}
 Claude-Sonnet-4.6$_\texttt{CoT}$       & \textbf{73.11} & 	14.81$^{\dagger}$ & 32.27$^{\dagger}$ \\
 Gemini-3.1-flash-lite                & 44.88 & 23.47 & 32.31 \\
  \rowcolor{softpeach}
 Gemini-3.1-flash-lite$_\texttt{CoT}$   & 72.60 & 23.99 & 35.72 \\
 Qwen3-Max                            & 49.02 & 17.79 & 26.31 \\
  \rowcolor{softpeach}
 Qwen3-Max$_\texttt{CoT}$               & 72.08 & \textbf{35.99} & \textbf{45.51} \\
 
 \midrule
 \textit{\textbf{Open-source LLMs}} \\
 Qwen3-8B           & 25.65 &  9.26 & 13.77 \\
 \rowcolor{softblue}
 Qwen3-8B$_\texttt{SFT}$         & 53.62 & 10.97 & 16.60 \\
 \rowcolor{softblue}
 Qwen3-8B$_\texttt{SFT+GRPO}$     & 53.77 & 11.79 & 17.98 \\
 LLaMA-3.3-70B            & 40.04 & 11.77 & 20.47 \\
 \rowcolor{softblue}
 LLaMA-3.3-70B$_\texttt{SFT}$       & \textbf{57.55} & \textbf{19.42} & \textbf{29.04} \\
 
    \bottomrule
    \end{tabular}}
    \caption{Performance comparison of LLMs on OmanicBench.
    \textbf{Bold} marks the best result per metric. $^{\dagger}$: discuss in Appendix~\ref{Discussion}.}
    \label{main_result}
    \vspace{-4mm}
\end{table}

\subsection{Models and Setup}
To assess the quality and utility of Omanic, we evaluate a set of proprietary and open-source LLMs under multiple settings. For proprietary models (e.g., GPT-5~\cite{gpt5}), we evaluate both direct answering and CoT prompting. For open-source models (e.g., Qwen3-8B~\cite{qwen3}), we additionally fine-tune selected models on the Omanic training set via supervised fine-tuning (SFT) and GRPO-based~\cite{grpo} reinforcement learning. To examine whether training on OmanicSynth transfers to broader reasoning capabilities, we further evaluate the fine-tuned models across multiple reasoning benchmarks (e.g., MATH~\cite{MATH}). 
For evaluation on OmanicBench, we report three complementary metrics across two evaluation paradigms. For the multiple-choice question setting in Table~\ref{main_result}, we use Multiple-Choice Question(MCQ) accuracy ~\cite{xinjie-etal-2025-reagent}, which measures selection accuracy over four candidate options. For the open-ended generation setting, we report Exact Match (EM), which requires strict string equivalence with the gold answer, and F1-Score, which captures partial credit at the token level~\cite{SQuAD}. 
We further report accuracy on each decomposed single-hop step in Table~\ref{step-level}, where each sub-question incorporates answers from preceding steps, enabling step-wise diagnosis beyond the final answer.

\begin{table}[t]
    \centering
    \scriptsize
    \resizebox{\columnwidth}{!}{
    \begin{tabular}{lrrrr}
    \toprule
     & \textbf{Step 1} & \textbf{Step 2} & \textbf{Step 3} & \textbf{Step 4} \\
    \midrule
    \textit{\textbf{Direct}} \\
    GPT-5.4 & 85.63 & 86.97 & 84.80 & \textcolor{red}{66.08} \\
    Claude-Sonnet-4.6 & 90.59 & 88.31 & 87.49 & \textcolor{red}{68.46} \\
    Gemini-3.1-Flash-Lite & 89.35 & 87.38 & 86.04 & \textcolor{red}{66.18} \\
    \midrule
    \textit{\textbf{CoT}} \\
    GPT-5.4 & 93.80 & 95.66 & 92.66 & \textcolor{red}{79.63} \\
    Claude-Sonnet-4.6 & 93.59 & 95.86 & 93.38 & \textcolor{red}{80.14} \\
    Gemini-3.1-Flash-Lite & 94.42 & 96.28 & 91.73 & \textcolor{red}{78.39} \\
    \bottomrule
    \end{tabular}}
    \caption{Single-hop step-level MCQ accuracy of selected models on OmanicBench. \textcolor{red}{Red} values indicate the lowest result for each model.}
    \label{step-level}
\end{table}

\subsection{Main Results}

Table~\ref{main_result} presents the overall performance of all evaluated models on OmanicBench. Proprietary LLMs consistently outperform open-source counterparts, while CoT prompting and OmanicSynth training both improve performance. For example, Qwen3-8B improves substantially after training on OmanicSynth (MCQ: 25.65 $\rightarrow$ 53.77), confirming that OmanicBench is challenging yet amenable to targeted training.
While Table~\ref{main_result} reports end-to-end performance, Table~\ref{step-level} highlights the diagnostic value of OmanicBench by exposing substantial variation across reasoning hops. Across all selected models and prompting settings, Step~4 is consistently the most difficult, with accuracy dropping far below earlier steps even when CoT is used. This pattern suggests that later-stage reasoning is not merely harder because of weaker models, but reflects an inherent bottleneck in composing multiple intermediate facts. 
Therefore, OmanicBench enables evaluation beyond final-answer correctness by localizing failures along the reasoning chain.
Notably, we also compare output length in Table~\ref{tab:output-length}: the gains from CoT for some models come at substantially higher inference costs. While Qwen3-Max$_\texttt{CoT}$ achieves the best open-ended performance, its output token length far exceeds that of all other models. In contrast, GPT-5.4$_\texttt{CoT}$ is considerably more efficient in practice. 
Full results including step-level accuracy in Appendix~\ref{sec:full-results}.


\begin{figure}[t]
    \centering
    \includegraphics[width=\linewidth]{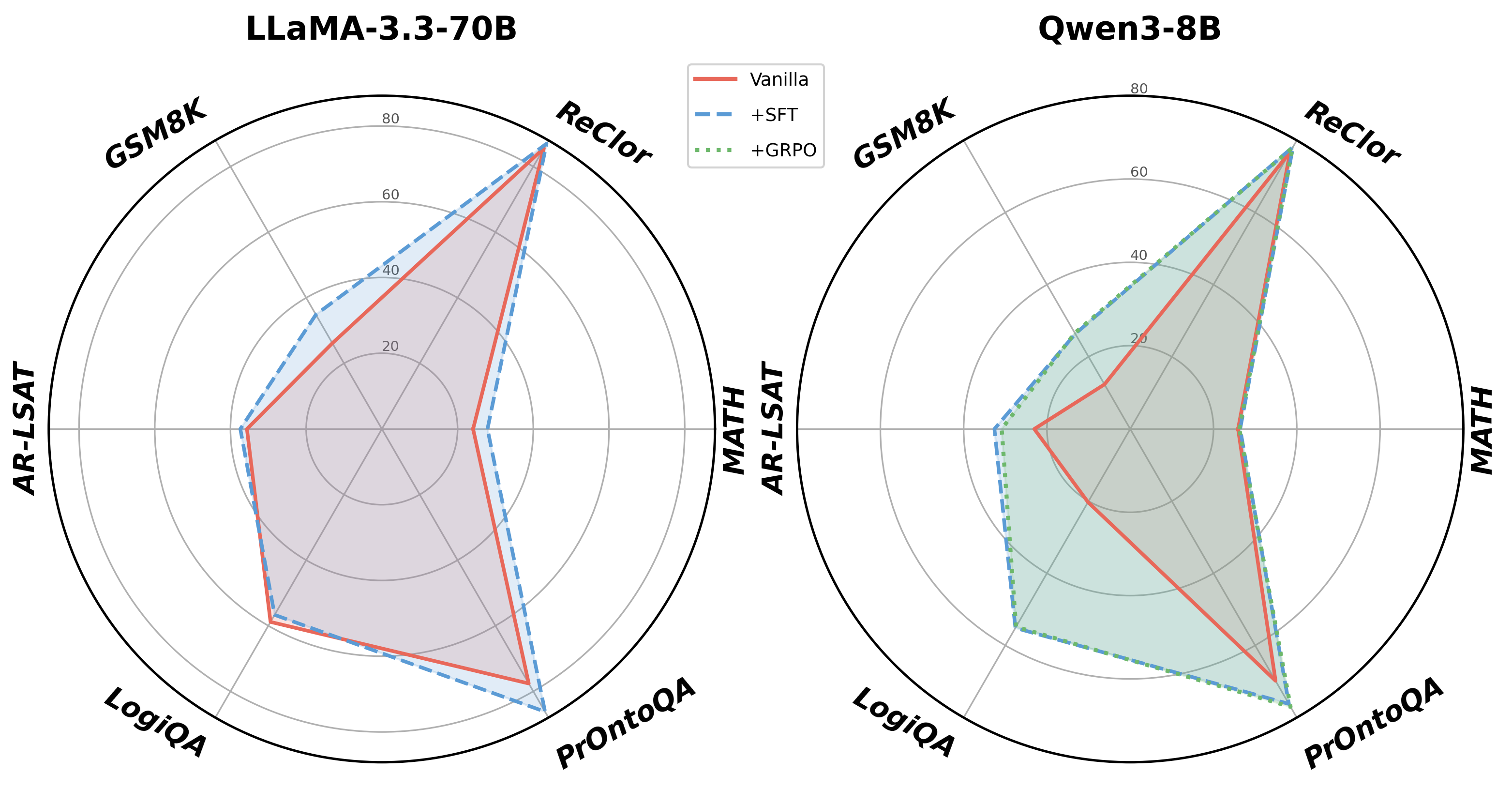}
    \caption{Comparison of accuracy across benchmarks between the vanilla model and its counterparts fine-tuned on OmanicSynth.}
    \label{benchmark}
    \vspace{-4mm}
\end{figure}

To further assess the transferability of skills acquired from OmanicSynth, we evaluate the fine-tuned models on established reasoning~\cite{ReClor,LogiQA,PrOntoQA} and mathematics~\cite{gsm8k} benchmarks. As shown in Figure~\ref{benchmark}, fine-tuned models consistently outperform their vanilla counterparts, demonstrating that OmanicSynth comprises high-quality, non-trivial instances that cultivate genuine complex logical reasoning and mathematical reasoning capabilities rather than superficial pattern matching or factual knowledge retrieval alone. All implementation details are provided in the Appendix~\ref{Implementation_details}.


\subsection{Key Observations}

Beyond aggregate performance, the step-level annotations in OmanicBench allow us to quantitatively examine three research questions about multi-hop reasoning behavior.

\begin{figure}[t]
    \centering
    \includegraphics[width=\linewidth]{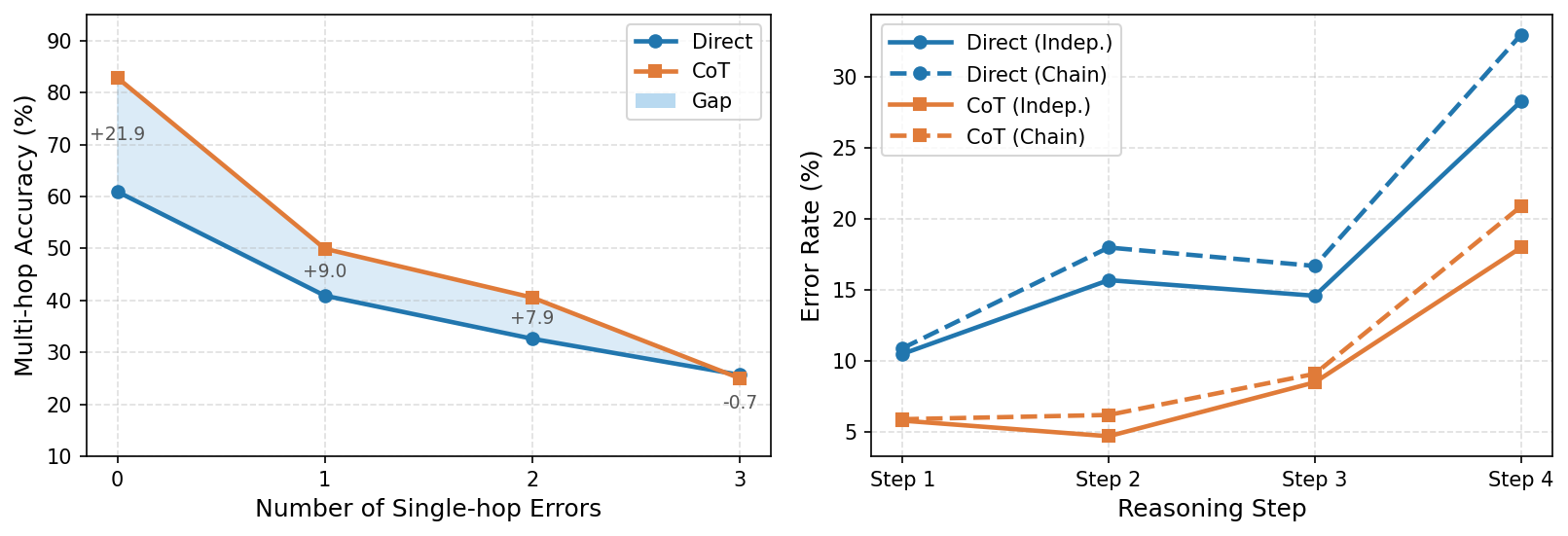}
    \caption{
    Results averaged across LLMs under Direct and CoT prompting.
Left: Multi-hop accuracy by number of single-hop errors. Right: Step-wise error rates under independent and chain evaluation. }
    \label{combined-analysis} 
    \vspace{-3mm}
\end{figure}

\textbf{RQ1: To what extent does CoT rely on a sufficient knowledge foundation?~\cite{cot_improve}}
To investigate this question, we group multi-hop questions by the number of constituent single-hop questions answered incorrectly, and then measure multi-hop accuracy within each group.
As shown in Figure~\ref{combined-analysis} (left), even in the zero-error group, multi-hop accuracy under Direct prompting reaches only about 60\%, well below ceiling, indicating that multi-hop reasoning and single-hop knowledge retrieval are not equivalent.
Meanwhile, CoT gain decreases monotonically as the number of erroneous single-hop steps increases, dropping to near zero ($-$0.7) when three steps are incorrect, while the largest gain (+21.9) is observed in the zero-error group.
These results quantify a clear knowledge floor on OmanicBench: effective CoT requires sufficient factual grounding, sharpening compositional inference but not substituting for missing facts.

\textbf{RQ2: To what extent do errors amplify toward the end of a multi-hop reasoning chain?~\cite{cot_gap}}
To quantify this effect, we compare two evaluation protocols: independent evaluation, where each step receives the gold answer to prior single-hop questions; and chain evaluation, where answers from previous steps propagate to subsequent steps.
As shown in Figure~\ref{combined-analysis} (right), even under independent evaluation, Step~4 already exhibits a substantially higher error rate than earlier steps, revealing an inherent difficulty gradient in OmanicBench beyond pure propagation effects.
Under chain evaluation, errors compound: Step~4 reaches a 33.0\% error rate under Direct prompting, 4.7 points higher than under independent evaluation.
While CoT reduces absolute error rates at every step, the amplification pattern remains intact, suggesting that sequential multi-hop inference is intrinsically more fragile in later hops.
Taken together, these analyses show that OmanicBench is not only a benchmark for end-to-end accuracy, but also a diagnostic testbed for quantifying where reasoning breaks down and how strongly errors compound across hops.

\textbf{RQ3: Does fine-tuning on OmanicSynth mainly inject factual knowledge, or does it teach models to reason across hops?}
To understand why fine-tuning on OmanicSynth improves multi-hop performance, we compare the answer entropy of vanilla Qwen3-8B and its fine-tuned variant Qwen3-8B$_\texttt{SFT}$ on OmanicBench. The key question is whether SFT primarily injects factual knowledge~(i.e., the model knows more) or teaches the model to leverage prior-hop information to progressively narrow the answer space~(i.e., the model reasons better). 
Lower entropy indicates higher confidence in the answer. Higher entropy indicates greater uncertainty, with probability mass spread across multiple options. 

\begin{figure}[t]
    \centering
    \includegraphics[width=\linewidth]{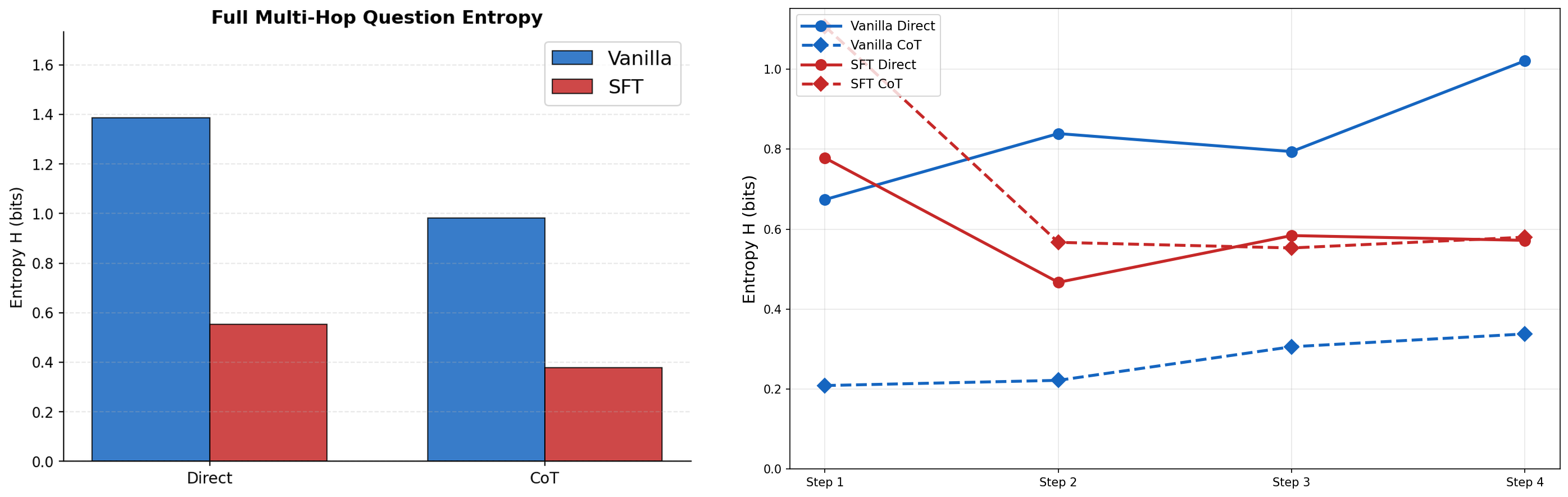}
    \caption{Answer entropy analysis for vanilla and SFT model. Left: Final answer entropy on full multi-hop questions. Right: Per-hop answer entropy under gold-context evaluation.}
    \label{Entropy_combined}
\end{figure}

When evaluated on full multi-hop questions, SFT substantially reduces final-answer entropy under both Direct and CoT prompting, as shown in Figure~\ref{Entropy_combined} (left). This indicates that the fine-tuned model produces more concentrated answer distributions regardless of whether step-by-step reasoning is explicitly elicited.
Notably, the entropy of the vanilla model under CoT remains higher than that of the SFT model under Direct prompting, suggesting that fine-tuning on OmanicSynth provides greater uncertainty reduction than explicit chain-of-thought guidance alone in this setting.

To move beyond end-to-end multi-hop performance, we further analyze answer entropy at the level of decomposed single-hop questions.
We analyze gold-context evaluation, where the model answers each hop with the full multi-hop question and the gold answers to all preceding hops. This setting tests whether the model can use correct prior-hop information to continue the reasoning chain and narrow the candidate answer space.
As shown in Figure~\ref{Entropy_combined} (right), the key signal is not that gold context always lowers entropy relative to the single-hop setting; in fact, adding prior-hop answers can slightly increase absolute entropy, suggesting that additional context may introduce distractors as well as useful evidence. Instead, the diagnostic signal lies in the direction of entropy change as gold information accumulates along the chain. Under Direct prompting, the vanilla model's entropy rises, indicating that it becomes increasingly uncertain as more context is provided. In contrast, the SFT model exhibits the opposite trajectory: entropy decreases even without explicit step-by-step prompting. This divergent trend, with rising entropy for vanilla but falling entropy for SFT under the same gold-context condition, shows that SFT teaches the model to use prior-hop information to progressively narrow the answer space rather than merely recall more facts.
Interestingly, vanilla Qwen3-8B under CoT exhibits even lower entropy than the SFT model. This does not contradict the effect of SFT; rather, it suggests that the relevant knowledge is largely present in the base model and can be elicited by an explicit reasoning scaffold. SFT therefore appears to internalize part of this scaffold, improving Direct prompting without requiring step-by-step instructions, while explicit CoT remains a strong external mechanism for organizing the model's latent knowledge.
Taken together, these entropy patterns suggest that fine-tuning on OmanicSynth primarily teaches the model to organize and propagate information across hops, rather than simply injecting additional factual knowledge.

We then analyze single-hop evaluation, where each decomposed sub-question is answered separately without the full multi-hop context or any preceding-hop answers. This setting serves as a step-level baseline for measuring the intrinsic uncertainty of individual hops.
As shown in Figure~\ref{Entropy_combined}, under Direct prompting, the entropy gap between vanilla and SFT is small at Hop~1 but grows substantially by Hop~4, indicating that the benefit of SFT becomes larger on later, more compositionally demanding steps. If SFT mainly injected missing factual knowledge, this gap would be expected to remain more uniform across hops. The growing gap instead suggests that SFT improves the model's ability to identify the relevant evidence and resolve deeper sub-questions. Moreover, vanilla Qwen3-8B under CoT attains entropy levels comparable to, and in some hops lower than, SFT under Direct prompting. Since CoT changes only the prompt rather than the model parameters, this pattern further indicates that the relevant knowledge is largely present in the base model, while SFT primarily teaches a more effective reasoning framework for extracting and organizing that knowledge.

\section{Conclusion}


Omanic is an open-domain 4-hop QA benchmark combining mathematical and factual reasoning with step-level annotations for diagnosing model failures.
Experiments show that CoT gains depend on factual completeness and errors accumulate across hops.
Together with OmanicSynth, it supports fine-grained evaluation and targeted training of multi-hop reasoning.


\section*{Limitations}
Omanic has several limitations that suggest directions for future work. First, the benchmark is restricted to English only, limiting its applicability to multilingual reasoning evaluation. Second, while 4-hop questions represent a meaningful increase in complexity over existing 2-hop benchmarks, extending to longer reasoning chains (e.g., 6-hop or 8-hop) would further test the limits of compositional reasoning. Third, although Omanic spans eight knowledge domains, certain specialized domains (e.g., legal, biomedical) remain underrepresented. Fourth, the dataset scale (10,296 training and 967 test instances) is moderate; scaling up through broader knowledge graph coverage could improve both training utility and evaluation robustness.
The motivation statement on Omanic can be found in Appendix~\ref{Statement}.

\section*{Acknowledgments}
Dr. Irene Li is supported by JST ACT-X (Grant
JPMJAX24CU) and JSPS KAKENHI (Grant
24K20832). This work used supercomputers provided by the Research Institute for Information
Technology, Kyushu University, through the HPCI
System Research Project (Project ID: hp260013).
\bibliography{custom}

\clearpage

\appendix

\section{Statement on OmanicBench Design and Scope}
OmanicBench is designed as a controlled diagnostic benchmark for studying multi-step reasoning in LLMs, rather than a new reasoning evaluation framework
Its fixed 4-hop structure allows us to compare models, localize reasoning failures, and quantify error propagation under different reasoning topologies.
These questions test abilities that are broadly useful in complex reasoning settings, including factual retrieval, mathematical reasoning, and compositional dependency tracking.
\label{Statement}

\section{Human Annotation Guidance \& Dataset Statistics}
\label{Guidance}
We recruit 10 undergraduate and graduate students to conduct the human correction and scoring process. 
To ensure ethical research practices, all annotators are compensated at a rate exceeding the local minimum wage. 
The entire project encompasses a cumulative total of 300 person-hours.
The detailed human annotation guidance is listed in the following:

\noindent For single-hop questions:

\begin{itemize}[leftmargin=15pt]
    \item \textbf{Factuality and Accuracy}: This criterion assesses the truthfulness and precision of the question-answer pair within real-world contexts, specific subject areas, or the provided knowledge source. Annotators follow four steps:
    \begin{enumerate}[leftmargin=15pt,nosep]
        \item \textit{Core Element Extraction.} Identify key entities (names, locations, events, dates) and logical relationships in the question.
        \item \textit{Cross-Source Verification.} Verify each entity's attributes using reliable databases, encyclopedias, or authoritative references.
        \item \textit{Terminology and Value Audit.} Check that specialized terms are spelled correctly and that any numerical computations are error-free.
        \item \textit{Final Scoring.} Assign a score based on the number and severity of errors (critical vs.\ minor).
    \end{enumerate}
    \begin{itemize}[leftmargin=10pt]
        \item \textbf{5 (Excellent)}: All key terms, concepts, dates, and numerical values are entirely accurate. The content is supported by authoritative evidence, remains free of "hallucinations" or misleading information, and utilizes precise terminology consistent with field conventions.
        \item \textbf{4 (Good)}: Most facts are accurate, with only minor flaws in non-core details that do not hinder overall understanding. Terminology is generally correct, though it may be slightly simplified while remaining acceptable within the domain.
        \item \textbf{3 (Satisfactory)}: Core facts are fundamentally correct, but errors exist that could lead to partial misunderstanding. Some key terms may be poorly translated or expressed, requiring the reader to infer the true intent based on common sense.
        \item \textbf{2 (Poor)}: Multiple critical facts or numerical values are incorrect, significantly impacting the understanding of the problem. Evident misuse of terminology or factual conflicts severely damages the professionalism of the content.
        \item \textbf{1 (Very Poor)}: Contains severe factual errors, false statements, or entirely fabricated "hallucinations." The question and answer fail to correspond, or most content contradicts known facts.
    \end{itemize}

    \item \textbf{Distractor Quality}: This criterion evaluates whether the three incorrect options (distractors) are sufficiently deceptive and belong to a plausible logical category. Annotators follow four steps:
    \begin{enumerate}[leftmargin=15pt,nosep]
        \item \textit{Category Consistency Check.} Verify that all distractors belong to the same logical category as the correct answer (e.g., if the answer is a location, distractors must also be locations).
        \item \textit{Domain Relevance Analysis.} Assess whether distractors share geographic, temporal, or thematic proximity with the correct answer within the same domain.
        \item \textit{Trap Logic Identification.} Examine whether distractors exploit plausible intermediate-step errors or common over-simplifications (e.g., arithmetic near-misses or temporally adjacent entities).
        \item \textit{Final Scoring.} Assign a score: if distractors are nearly indistinguishable without full reasoning, score 5; if they span unrelated categories, score 1--2.
    \end{enumerate}
    \begin{itemize}[leftmargin=10pt]
        \item \textbf{5 (Highly Deceptive)}: All distractors belong to the same domain. Numerical options represent logical "traps" (e.g., calculation near-misses), while semantic options consist of easily confused synonyms, similar institutions, or entities derived from intermediate reasoning steps.
        \item \textbf{4 (Effective)}: Distractors are within a reasonable scope (e.g., for a question about Europe, all distractors are European). The format is highly consistent with the answer, making them impossible to exclude via simple word-class or logical loopholes.
        \item \textbf{3 (Moderate)}: Distractors share attributes with the answer (e.g., both are years or locations), but one or two options can be quickly excluded using general knowledge or obvious context clues.
        \item \textbf{2 (Weak)}: Distractors belong to the same broad category but possess obvious attribute differences (e.g., asking for a university name while providing a country name as a distractor) or inconsistent formatting.
        \item \textbf{1 (Non-existent)}: Distractors are completely irrelevant or cross-domain (e.g., a "year" question with "apple" as an option). These are illogical "fillers" that can be instantly identified.
    \end{itemize}

    \item \textbf{Fluency and Completeness}: This criterion evaluates whether the language is natural, the logic is clear, the grammar is correct, and all necessary constraints for a unique answer are provided. Annotators follow four steps:
    \begin{enumerate}[leftmargin=15pt,nosep]
        \item \textit{Naturalness First-Read.} Read the question as a native speaker, checking for awkward phrasing, inverted word order, or logical discontinuities---paying special attention to subordinate clauses and pronoun references in longer sentences.
        \item \textit{Constraint Completeness Scan.} Verify that the question stem contains every constraint necessary to derive a unique correct answer (e.g., specific years, inclusive/exclusive conditions).
        \item \textit{Grammar and Spelling Check.} Inspect punctuation, capitalization of proper nouns, and tense consistency.
        \item \textit{Final Scoring.} Assign a score: flawless grammar with a self-contained information loop scores 5; translation artifacts or missing critical constraints lower the score accordingly.
    \end{enumerate}
    \begin{itemize}[leftmargin=10pt]
        \item \textbf{5 (Excellent)}: Expression is extremely natural and smooth, fully adhering to native usage habits. The structure is rigorous with appropriate logical connectives and zero grammatical errors. All original details and necessary constraints are preserved without omission.
        \item \textbf{4 (Good)}: Expression is basically natural with only minor linguistic flaws. The main meaning is preserved, though some non-critical descriptive details may be missing. Sentence structures follow standard norms with negligible errors.
        \item \textbf{3 (Satisfactory)}: Expression is slightly rigid or exhibits a "translation-ese" tone. While the core meaning is conveyed, it may omit some important constraints or include irrelevant information; grammar errors are present but the meaning remains clear.
        \item \textbf{2 (Poor)}: Lacks fluency with abrupt transitions and unclear logical connections. Core information is incomplete, containing significant omissions or serious errors that affect the ability to judge the correct answer.
        \item \textbf{1 (Very Poor)}: Language is extremely unnatural or unintelligible, containing severe grammatical and logical errors. The content organization is chaotic and fails to reflect the original intent of the question.
    \end{itemize}
\end{itemize}

\noindent For 4-hop questions:

\begin{itemize}[leftmargin=15pt]
    \item \textbf{Logical Integrity and Technical Formatting}: This criterion evaluates the structural rigor of the multi-hop reasoning chain and the standardized use of LaTeX, technical terminology, and code formatting. Annotators follow four steps:
    \begin{enumerate}[leftmargin=15pt,nosep]
        \item \textit{Logic Chain Decomposition.} Break the multi-hop question into an explicit path $A \rightarrow B \rightarrow C \rightarrow D$ and identify each intermediate node.
        \item \textit{Input--Output Matching.} Verify that the output of each preceding hop is correctly used as the input condition for the subsequent hop.
        \item \textit{Symbol and Code Audit.} Check that all LaTeX formulas, mathematical notation, currency symbols, and code blocks are correctly rendered and unaltered.
        \item \textit{Consistency Determination.} Confirm that the logic chain forms a closed loop with no breaks or circular reasoning.
    \end{enumerate}
    \begin{itemize}[leftmargin=10pt]
        \item \textbf{5 (Excellent)}: The reasoning chain is perfectly airtight without gaps or circularity. All LaTeX symbols, mathematical formulas, and currency signs (\$) are technically flawless and correctly rendered.
        \item \textbf{4 (Good)}: The logical chain is complete and sound, though the natural language transitions between hops may feel slightly rigid. Technical formatting is basically standardized with only negligible layout flaws.
        \item \textbf{3 (Satisfactory)}: A minor logical flaw exists (e.g., ambiguous pronoun reference like "the artist" when multiple artists are mentioned), requiring the reader to re-read to clarify the steps. Terminology or LaTeX symbols may show slight formatting deviations.
        \item \textbf{2 (Poor)}: Relationships between multiple hops are unclear, or critical premises are lost due to poor phrasing, preventing a successful logical "closed-loop." Formatting is chaotic, with LaTeX symbols or code blocks appearing garbled.
        \item \textbf{1 (Very Poor)}: The logical chain is entirely broken or results in a paradox (e.g., asking for a "year" but requiring a "monetary amount" as the answer). Terminology is highly unprofessional, and formatting has completely collapsed.
    \end{itemize}

    \item \textbf{Contextual Fact Consistency}: This criterion verifies whether all single-hop facts remain factually accurate and conflict-free when amalgamated into a multi-hop narrative. Annotators follow four steps:
    \begin{enumerate}[leftmargin=15pt,nosep]
        \item \textit{Atomic Fact Backtracking.} Compare each background claim in the multi-hop question (e.g., a date, a title, a numeric value) against the original single-hop data.
        \item \textit{Spatiotemporal Conflict Detection.} Verify that the combined timeline is logically coherent (e.g., an appointment in 1877 requires the predecessor's tenure to overlap or precede that date).
        \item \textit{Modifier Verification.} Check whether qualifiers added during composition (e.g., ``the large art school'') inadvertently alter the original meaning.
        \item \textit{Final Scoring.} If all intermediate-node facts are correct and transitions are natural, assign the full score.
    \end{enumerate}
    \begin{itemize}[leftmargin=10pt]
        \item \textbf{5 (Excellent)}: All integrated facts (dates, locations, values) are logically consistent with one another and the real-world background. The combined scenario is realistic (e.g., a museum established in 2000 is correctly described as existing in 2005).
        \item \textbf{4 (Good)}: Core facts are accurate, but minor descriptive biases in non-essential details (e.g., secondary institutional titles or honorifics) occur during amalgamation without affecting the final answer.
        \item \textbf{3 (Satisfactory)}: Core facts remain fundamentally correct, but the timeline or logical background across hops feels slightly "awkward" or strained, though no direct factual conflict is present.
        \item \textbf{2 (Poor)}: Significant factual flaws appear in intermediate steps (e.g., a single-hop answer is "11 years" but is treated as "12 years" during the multi-hop calculation), rendering the final result unreliable.
        \item \textbf{1 (Very Poor)}: A severe factual error exists in at least one intermediate step, or the timeline is logically impossible (e.g., a divorce occurring before the marriage).
    \end{itemize}

    \item \textbf{Answer Obscurity and Leakage Prevention}: This criterion evaluates whether the question stem accidentally reveals the final answer or if the reasoning steps provide a sufficient challenge. Annotators follow four steps:
    \begin{enumerate}[leftmargin=15pt,nosep]
        \item \textit{Keyword Filtering.} Search the question stem for the final answer itself or any strongly characteristic cues that directly point to it.
        \item \textit{Shortcut Test.} Attempt to reach the correct answer without completing the intermediate hops---relying only on the final segment of the question or general knowledge.
        \item \textit{Distractor Elimination Check.} Assess whether the stem provides enough non-logical information to rule out all incorrect options without genuine reasoning.
        \item \textit{Final Scoring.} If every reasoning step is indispensable for reaching the answer, score 5; if the final segment alone makes the answer obvious, lower the score accordingly.
    \end{enumerate}
    \begin{itemize}[leftmargin=10pt]
        \item \textbf{5 (Excellent)}: No leakage. The solver must complete every reasoning step to find the answer. The final answer or its distinct characteristics do not appear, directly or indirectly, within the question stem.
        \item \textbf{4 (Good)}: The reasoning chain is intact, though some background descriptions might allow a model to narrow down the answer range via a process of elimination rather than pure deduction.
        \item \textbf{3 (Satisfactory)}: Partial leakage occurs. Certain phrasing is too direct (e.g., mentioning a highly unique year or rare proper noun), allowing experienced annotators or models to "guess" the answer via shortcuts or common sense.
        \item \textbf{2 (Poor)}: The question contains obvious hints that make the reasoning chain significantly easier to bypass.
        \item \textbf{1 (Very Poor)}: Direct leakage. The final answer appears within the multi-hop description, or the question is phrased in a way that renders the logical steps meaningless.
    \end{itemize}

    \item \textbf{Semantic Completeness and Linguistic Fluency}: This criterion evaluates the retention of all necessary constraints and the naturalness of the linguistic expression. Annotators follow four steps:
    \begin{enumerate}[leftmargin=15pt,nosep]
        \item \textit{Grammar and Rhetoric Scan.} Check long, complex sentences for grammatical errors and ambiguous references (e.g., multiple uses of ``the artist'' when several artists are mentioned).
        \item \textit{Constraint Condition Checklist.} Confirm that all critical constraints from the single-hop questions (e.g., ``since 1855,'' ``prior to'') are faithfully carried over into the multi-hop question.
        \item \textit{Logical Connector Check.} Verify that connectors such as ``prior to,'' ``who,'' and ``where'' accurately reflect the inter-hop relationships.
        \item \textit{Final Scoring.} A question that reads fluently and preserves all constraints scores 5; noticeable ``translation-ese'' or ambiguous references lower the score to 3 or below.
    \end{enumerate}
    \begin{itemize}[leftmargin=10pt]
        \item \textbf{5 (Excellent)}: All necessary constraints (e.g., specific year ranges, "inclusive," rounding requirements) are perfectly preserved. The language is natural, smooth, and adheres to native-speaker habits with precise logical connectives.
        \item \textbf{4 (Good)}: Constraints are complete, but the phrasing is slightly wordy. Language is fluent, though the choice of logical connectors (e.g., "prior to," "who") may be repetitive.
        \item \textbf{3 (Satisfactory)}: The core meaning is clear, but some non-essential constraints are omitted (e.g., missing a rounding instruction). The expression is rigid and exhibits "translation-ese" or a formulaic tone.
        \item \textbf{2 (Poor)}: Critical constraints required for derivation are missing, or redundant information is added that interferes with understanding. Logical connectors are used incorrectly.
        \item \textbf{1 (Very Poor)}: Key constraints are missing, making it impossible to determine a unique answer. The language is broken and the logical relationships are erroneous, making the text unreadable.
    \end{itemize}

    \item \textbf{Reasoning Complexity and Domain Diversity}: This criterion measures the degree of cross-domain complexity and the depth of the logical jumps. Annotators follow four steps:
    \begin{enumerate}[leftmargin=15pt,nosep]
        \item \textit{Domain Counting.} Identify the number of distinct knowledge domains spanned by the question (e.g., Literature, Geography, Art History, Arithmetic).
        \item \textit{Hop Counting.} Count the number of explicit logical transitions from the starting entity to the final answer.
        \item \textit{Depth and Dependency Analysis.} Determine whether each hop requires domain-specific knowledge that cannot be bypassed through common sense alone.
        \item \textit{Level Determination.} Assign a score based on the number of domains crossed (4+ domains = top tier) and the number of non-trivial hops.
    \end{enumerate}
    \begin{itemize}[leftmargin=10pt]
        \item \textbf{5 (Excellent)}: The logical chain spans 4 or more distinct domains (e.g., Art History $\rightarrow$ Geography $\rightarrow$ Law $\rightarrow$ Financial Arithmetic). Each step is strictly dependent on the previous one and cannot be bypassed via common sense.
        \item \textbf{4 (Good)}: The logic spans 3 distinct domains and involves at least 4 explicit logical jumps.
        \item \textbf{3 (Satisfactory)}: The reasoning chain involves 3 or more deep logical steps within at least 2 distinct domains. It effectively distinguishes models with single-domain knowledge.
        \item \textbf{2 (Poor)}: The chain involves 2 domains but only 1--2 simple logical jumps, or one of the domains relies on common knowledge.
        \item \textbf{1 (Very Poor)}: "Pseudo-multi-hop." The logic is extremely simple, consisting merely of the additive stacking of facts within the same domain.
    \end{itemize}
\end{itemize}

\begin{figure}[t]
    \centering
    \includegraphics[width=0.8\linewidth]{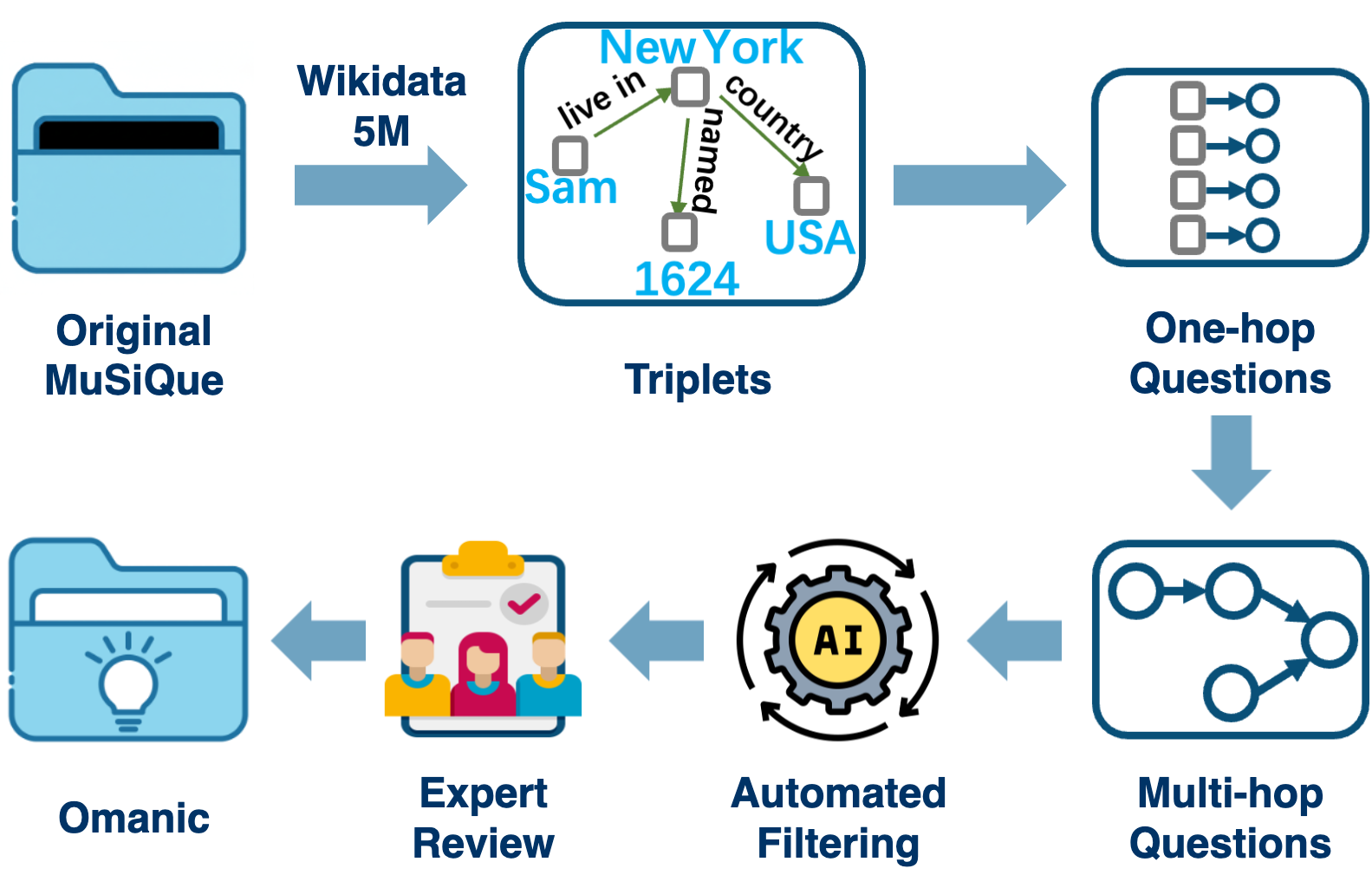}
    \caption{Overview of Omanic construction pipeline, where rectangles denote entities and circles denote single-hop questions.}    \label{pipeline}
    \vspace{-4mm}
\end{figure}

\begin{figure*}[t] 
    \centering
    \includegraphics[width=1.0\textwidth]{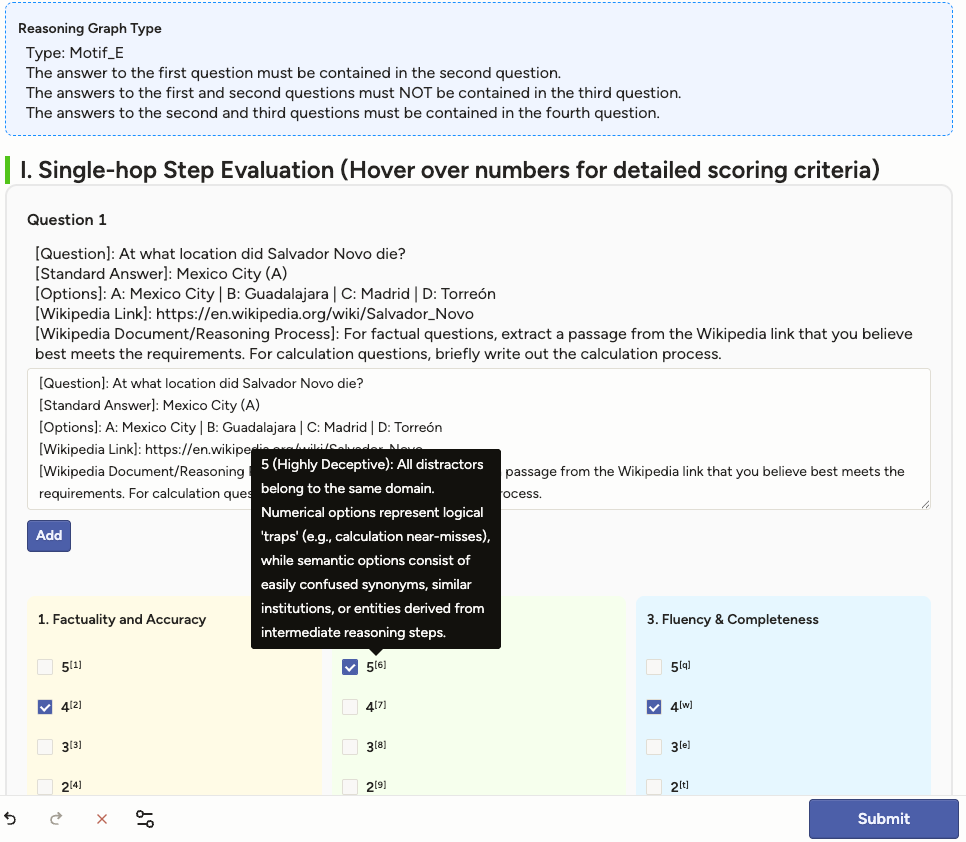} 
    \caption{Annotation interface for single-hop question.}
    \label{single_hop}
\end{figure*}

\begin{figure*}[t] 
    \centering
    \includegraphics[width=1.0\textwidth]{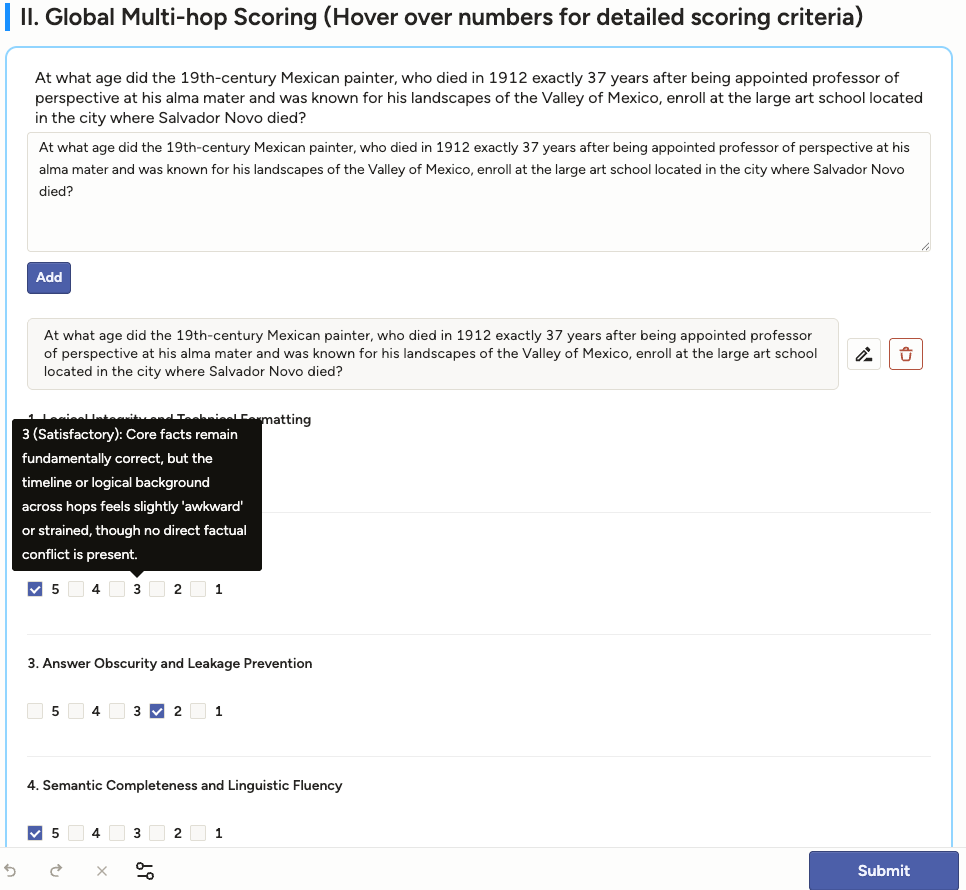} 
    \caption{Annotation interface for 4-hop question.}
    \label{4-hop}
\end{figure*}

\begin{table}[t]
    \centering
    \begin{tabular}{lccc}
    \toprule
    \multicolumn{4}{c}{\textit{Single-hop Sub-questions}} \\
    \midrule
     & \textbf{Factuality} & \textbf{Distractor} & \textbf{Fluency} \\
    \midrule
    Mean$_\text{STD}$ & $4.62_{0.54}$ & $4.31_{0.72}$ & $4.47_{0.61}$ \\
    \midrule
    \multicolumn{4}{c}{\textit{4-hop Questions}} \\
    \midrule
     & \textbf{Logic} & \textbf{Consistency} & \textbf{Obscurity} \\
    \midrule
    Mean$_\text{STD}$ & $4.53_{0.58}$ & $4.41_{0.65}$ & $4.18_{0.79}$ \\
    \midrule
     & \textbf{Fluency} & \textbf{Complexity} & \\
    \midrule
    Mean$_\text{STD}$ & $4.44_{0.62}$ & $4.56_{0.53}$ & \\
    \bottomrule
    \end{tabular}
    \caption{Human annotation scores on a 1--5 scale over 967 test instances. Each cell reports the mean score with the standard deviation in subscript (mean$_\text{std}$).}
    \label{tab:annotation-scores}
\end{table}

\clearpage


\renewcommand{\arraystretch}{1.25}

\begin{table*}[t]
\centering
\scriptsize
\setlength{\tabcolsep}{8pt}

\begin{tabularx}{\textwidth}{
  >{\centering\arraybackslash}p{0.18\textwidth}
  p{0.28\textwidth}
  Y
}

\hline
\textbf{Reasoning Graph} & \textbf{Multi-hop Question} & \textbf{Single-hop Composition} \\
\hline

\vspace{0pt}{\includegraphics[width=0.18\textwidth,keepaspectratio]{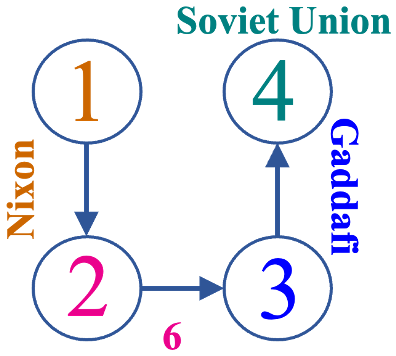}}

&
\vspace{0pt}
If a political regime began in 1969 and lasted for a number of years equal to 7 times the count of Apollo moon landing missions that occurred during the presidency of Eisenhower's vice president, and if the U.S. was one of 2 major powers that recognized this regime's leader early on, what was the other major power?
\textcolor{teal}{Soviet Union}

\noindent\textit{[A: United Kingdom B: France C: Soviet Union D: West Germany]}

&
\begin{enumerate}[leftmargin=*,nosep,topsep=0pt,partopsep=0pt]
    \item Who served as Eisenhower's vice president?
    \textcolor{orange!80!black}{Nixon}
    \item How many successful Apollo moon landing missions occurred while \textcolor{orange!80!black}{Nixon} was president of the U.S.?
    \textcolor{magenta}{6}
    \item If a political regime lasted for \textcolor{magenta}{6} times 7 years starting from 1969, whose face was most closely associated with Libya's government during this period?
    \textcolor{blue}{Gaddafi}
    \item If the U.S. was one major power that recognized \textcolor{blue}{Gaddafi}'s government at an early date, and there were 2 major powers total that did so early on, what was the other major power besides the U.S.?
    \textcolor{teal}{Soviet Union}
\end{enumerate}
\\
\hline
\vspace{0pt}{\includegraphics[width=0.18\textwidth,keepaspectratio]{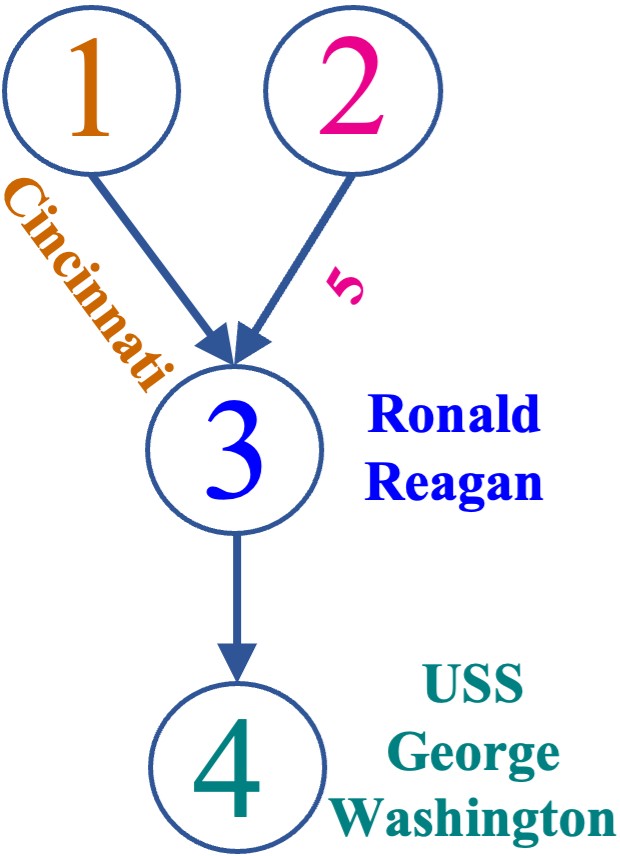}}

&
\vspace{0pt}
Great American Ball Park is the home stadium of an MLB team that has won the World Series multiple times. Multiply their total championship count by the number of players permanently banned for fixing the 1919 World Series in the Black Sox Scandal. The carrier named after the president with that ordinal number replaced which vessel at Naval Station Yokosuka, Japan, in 2015?
\textcolor{teal}{USS George Washington}

\noindent\textit{[A: USS Nimitz B: USS Carl Vinson C: USS George Washington D: USS John C. Stennis]}

&

\begin{enumerate}[leftmargin=*,nosep,topsep=0pt,partopsep=0pt]
    \item In which U.S. city is Great American Ball Park located?
    \textcolor{orange!80!black}{Cincinnati}
    \item How many World Series championships have the Cincinnati Reds won in total? 
    \textcolor{magenta}{5}
    \item Eight Chicago White Sox players were permanently banned from baseball for their role in fixing the 1919 World Series (the Black Sox Scandal). Multiply this number by the \textcolor{orange!80!black}{Cincinnati} Reds' total World Series championships (\textcolor{magenta}{5}). Which U.S. President held the ordinal number equal to this product?
    \textcolor{blue}{Ronald Reagan}
    \item USS \textcolor{blue}{Ronald Reagan} replaced which aircraft carrier as the U.S. Navy's forward-deployed vessel at Naval Station Yokosuka, Japan, in 2015?
    \textcolor{teal}{USS George Washington}
\end{enumerate}
\\
\hline

\end{tabularx}
\caption{Example multi-hop reasoning graph and its step-by-step question decomposition.}
\label{example_app}
\end{table*}

\begin{table*}[t]
    \centering
    \small
    \resizebox{\textwidth}{!}{
    \begin{tabular}{lcccccc}
    \toprule
    \textbf{Dataset} & \textbf{Open Domain} & \textbf{\# Hops} & \textbf{Explicit step-wise chain} & \textbf{Topology Annotated} & \textbf{Math reasoning} & \textbf{Expert-review} \\
    \midrule
    HotpotQA-SubQ~\cite{HotpotQA-SubQ} & Yes & 2 & Yes & No & No & No \\
    MuSiQue~\cite{musique} & Yes & 2-4 & No & No & No & Yes \\
    FanOutQA~\cite{fanoutqa} & Yes &  \textbf{5-6} & Yes & No  & No & Yes \\
    MoreHopQA~\cite{Morehopqa} & No & 1-5 & Yes & No & Yes & Yes \\
    CofCA~\cite{cf_multihop} & Yes & 2-4 & No & No & No & No \\
    SynthWorlds~\cite{synthworlds} & No & 2-4 & No & No & No & No \\
    \textbf{OmanicBench (ours)} & \textbf{Yes} & 4 & \textbf{Yes} & \textbf{Yes} & \textbf{Yes} & \textbf{Yes} \\
    \bottomrule
    \end{tabular}
    }
    \caption{Qualitative comparison between OmanicBench and representative multi-hop or reasoning benchmarks. ``\# Hops'' reports the supported hop range; for MuSiQue and CofCA, around 80\% of the questions are 2-hop. ``Explicit step-wise chain'' indicates whether a benchmark provides explicit step-level annotations, such as decomposed sub-questions and intermediate answers, for diagnosing reasoning failures. ``Topology Annotated'' indicates whether a benchmark explicitly defines and annotates reasoning topologies (e.g., chain, tree, DAG) for its questions.}
    \label{tab:benchmark-comparison}
\end{table*}

\clearpage

\begin{figure*}[t]
    \centering
    \begin{tabular}{c@{\hskip 0.8em}c@{\hskip 0.8em}c}
        \includegraphics[height=2.8cm]{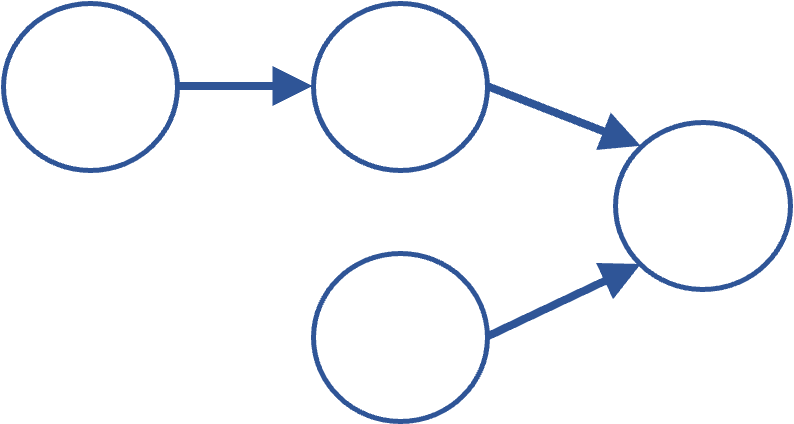} &
        \includegraphics[height=2.8cm]{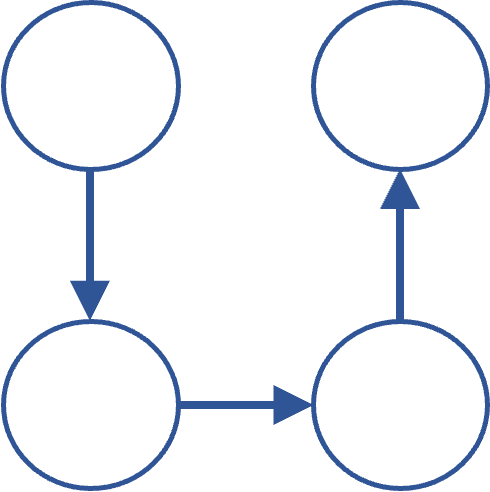} &
        \includegraphics[height=2.8cm]{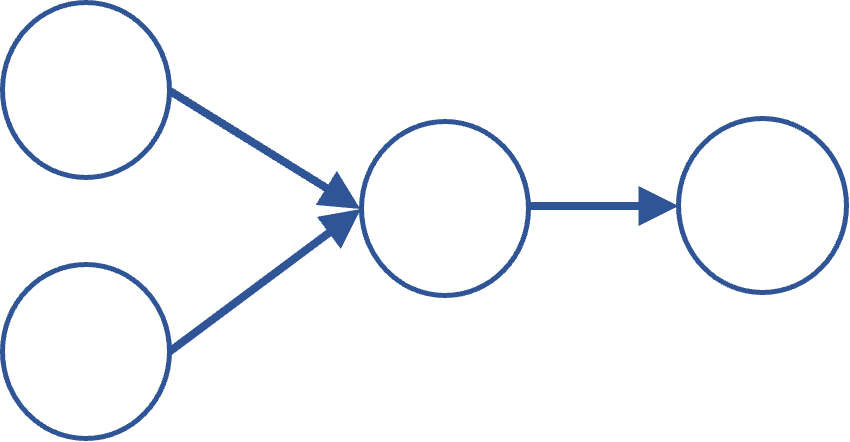} \\[6pt]
        (a) Bridge & (b) Chain & (c) Converging
    \end{tabular}
    \caption{Three reasoning graph topologies used in Omanic for organizing 4-hop questions.}
    \label{reasoning-graphs}
\end{figure*}

\begin{figure*}[t]
    \centering
    \includegraphics[width=\textwidth]{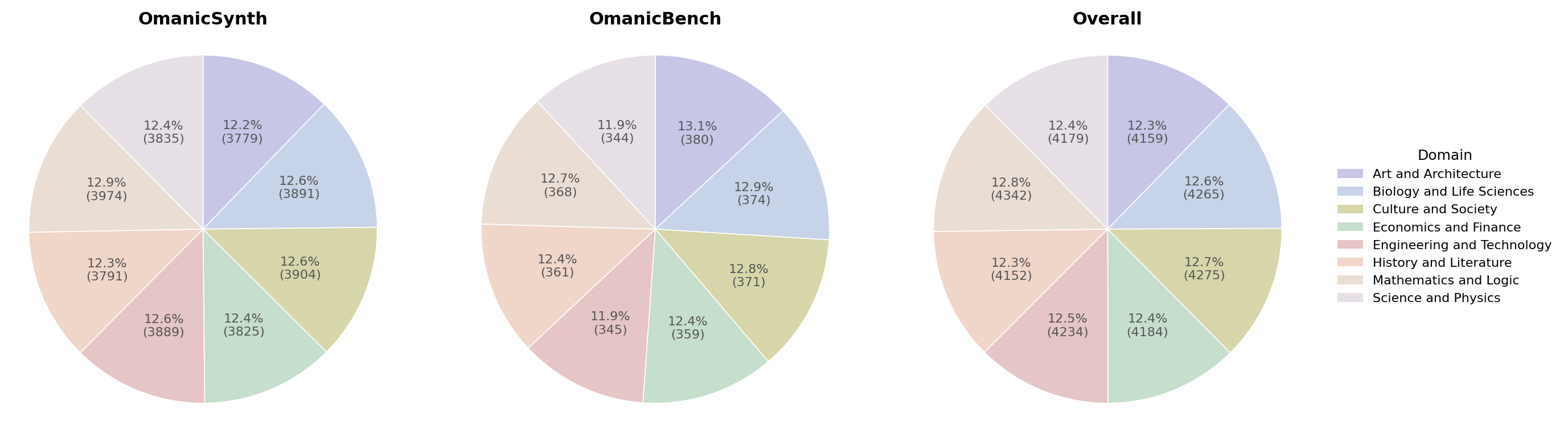}
    \caption{Domain distribution of single-hop sub-questions across OmanicSynth, OmanicBench, and the overall dataset.}
    \label{domain-distribution}
\end{figure*}

\begin{figure*}[t]
    \centering
    \includegraphics[width=\textwidth]{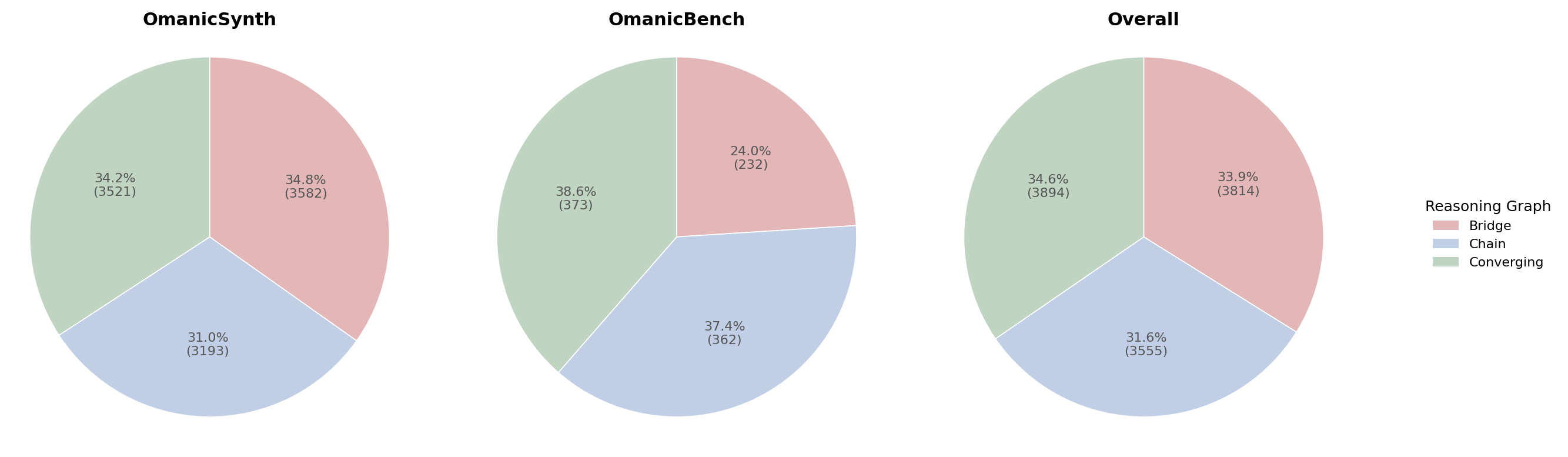}
    \caption{Reasoning graph topology distribution across OmanicSynth, OmanicBench, and the overall dataset.}
    \label{graph-topology-distribution}
\end{figure*}


\clearpage
\section{Full Results}
\label{sec:full-results}
\subsection{Implementation details}\label{Implementation_details}
For supervised fine-tuning, we construct a mixed training set in which half of the instances are formatted as multi-choice question examples and the other half are formatted as open-ended generation examples. For Qwen3-8B, we adopt a two-stage training framework: we first perform full-parameter SFT, and then conduct GRPO-based~\cite{grpo} reinforcement learning starting from the fine-tuned model. For LLaMA-3.3-70B, we perform LoRA-based~\cite{lora} SFT on the same mixed training data. All experiments are conducted on 4 NVIDIA 96GB H100 NVL GPUs, and the detailed training hyperparameters are reported in Tables~\ref{tab:qwen-sft-hparams}, \ref{tab:qwen-grpo-hparams}, and \ref{tab:llama-lora-hparams}. The evaluation templates are summarized in Table~\ref{tab:eval-templates}.

\begin{table}[h]
    \centering
    \small
    \begin{tabularx}{\columnwidth}{lY}
    \toprule
    \textbf{Setting} & \textbf{Template} \\
    \midrule
    Direct multi-choice question & Select the correct option from four candidates and return only the answer letter (A/B/C/D). \\
    Direct open-ended & Answer as concisely as possible and provide only the final answer without explanation. \\
    CoT multi-choice question & Think step by step and end the response with ``The answer is X'', where X is A, B, C, or D. \\
    CoT open-ended & Think step by step and write the final answer on a separate line in the format ``FINAL ANSWER: <answer>''. \\
    \bottomrule
    \end{tabularx}
    \caption{Evaluation templates for direct and CoT prompting.}
    \label{tab:eval-templates}
\end{table}

\begin{table}[h]
    \centering
    \small
    \begin{tabularx}{\columnwidth}{lY}
    \toprule
    \textbf{Parameter} & \textbf{Value} \\
    \midrule
    Cutoff length & 2048 \\
    Per-device batch size & 32 \\
    Gradient accumulation & 2 \\
    Learning rate & $1.0\times10^{-5}$ \\
    Epochs & 3.0 \\
    LR scheduler & Cosine \\
    Warmup ratio & 0.1 \\
    Precision & BF16 \\
    \bottomrule
    \end{tabularx}
    \caption{Hyperparameters for Qwen3-8B Full SFT.}
    \label{tab:qwen-sft-hparams}
\end{table}

\begin{table}[h]
    \centering
    \small
    \begin{tabularx}{\columnwidth}{lY}
    \toprule
    \textbf{Parameter} & \textbf{Value} \\
    \midrule
    Train batch size & 512 \\
    Max prompt length & 1024 \\
    Max response length & 1024 \\
    Actor learning rate & $1.0\times10^{-5}$ \\
    PPO mini-batch size & 512 \\
    PPO micro-batch size / GPU & 32 \\
    PPO epochs & 1 \\
    KL loss & Enabled \\
    KL coefficient & 0.01 \\
    Number of rollouts & 5 \\
    Total epochs & 3 \\
    \bottomrule
    \end{tabularx}
    \caption{Hyperparameters for Qwen3-8B GRPO training.}
    \label{tab:qwen-grpo-hparams}
\end{table}

\begin{table}[h]
    \centering
    \small
    \begin{tabularx}{\columnwidth}{lY}
    \toprule
    \textbf{Parameter} & \textbf{Value} \\
    \midrule
    LoRA rank & 8 \\
    Cutoff length & 2048 \\
    Per-device batch size & 32 \\
    Gradient accumulation & 2 \\
    Learning rate & $1.0\times10^{-4}$ \\
    Epochs & 3.0 \\
    Precision & BF16 \\
    \bottomrule
    \end{tabularx}
    \caption{Hyperparameters for LLaMA-3.3-70B LoRA SFT.}
    \label{tab:llama-lora-hparams}
\end{table}

\begin{table}[h]
    \centering
    \resizebox{\columnwidth}{!}{
    \begin{tabular}{lrr}
    \toprule
          & \textbf{Direct} & \textbf{CoT}  \\
    \midrule
    \textit{\textbf{OpenAI}} \\
    GPT-5.4                & 13    & 451   \\
    GPT-5.2                & 15    & 502   \\
    GPT-5.1                & 12    & 1,126 \\
    GPT-4o                 & 16    & 1,314 \\
    \midrule
    \textit{\textbf{Anthropic}} \\
    Claude-Sonnet-4.6      & 611   & 1,812 \\
    Claude-Sonnet-4.5      & 82    & 1,379 \\
    Claude-Opus-4.5        & 332   & 1,469 \\
    Claude-Opus-4.1        & 196   & 1,221 \\
    Claude-Sonnet-4        & 76    & 1,316 \\
    Claude-Opus-4          & 112   & 1,223 \\
    \midrule
    \textit{\textbf{Google}} \\
    Gemini-3.1-flash-Lite  & 10    & 1,423 \\
    Gemini-3-Flash-Preview & 9     & 1,587 \\
    Gemini-2.5-Flash       & 8     & 3,837 \\
    Gemini-2.5-Flash-Lite  & 8     & 6,453 \\
    \midrule
    \textit{\textbf{Meta}} \\
    LLaMA-3.3-70B          & 38    & 1,694 \\
    LLaMA-3-8B             & 22    & 1,469 \\
    LLaMA-3-70B            & 11    & 735   \\
    \midrule
    \textit{\textbf{Alibaba}} \\
    Qwen3-Max              & 14    & 4,841 \\
    Qwen3-32B              & 33    & 920   \\
    Qwen3-8B               & 24    & 357   \\
    Qwen2.5-72B            & 12    & 1,260 \\
    Qwen2.5-7B             & 15    & 1,313 \\
    \midrule
    \textit{\textbf{DeepSeek}} \\
    DeepSeek-V3.2                 & 11    & 2,548 \\
    DeepSeek-R1-Distill-LLaMA-70B & 111   & 652   \\
    DeepSeek-R1-Distill-Qwen-32B  & 68    & 429   \\
    \bottomrule
    \end{tabular}}
    \caption{Average output length under Direct and CoT prompting.}
    \label{tab:output-length}
\end{table}

\subsection{Discussion}\label{Discussion}
As marked by $^{\dagger}$ in Table~\ref{main_result}, Claude-Sonnet-4.6 exhibits a unique failure mode under CoT prompting: its MCQ accuracy improves substantially, yet open-ended performance (EM/F1) degrades. Manual inspection reveals that Claude's CoT responses are substantially longer (avg. 1,812 tokens vs. GPT-5.4's 451 tokens), with the final answer frequently buried in extended prose rather than presented in an extractable format.  This suggests the degradation reflects an answer extraction failure rather than a reasoning regression, underscoring the importance of evaluating reasoning capability (MCQ) and answer articulation (EM/F1) as complementary, not interchangeable, axes.

\clearpage

\begin{table}[t]
    \centering
    \resizebox{\columnwidth}{!}{
    \begin{tabular}{lccc}
    \toprule
         & \textbf{MCQ} & \textbf{Exact Match} & \textbf{F1-Score} \\
       \midrule
        \textit{\textbf{OpenAI}} \\
GPT-5.2                              & 38.51 & 18.20 & 29.09 \\
 \rowcolor{softpeach}
GPT-5.2$_\texttt{CoT}$               & 65.98 & 34.13 & 46.93 \\
GPT-5.1                              & 39.40 & 18.55 & 26.62 \\
 \rowcolor{softpeach}
GPT-5.1$_\texttt{CoT}$               & 71.68 & 39.96 & 51.18 \\
GPT-4o                               & 39.50 & 16.65 & 28.93 \\
 \rowcolor{softpeach}
GPT-4o$_\texttt{CoT}$                & 66.91 & 37.85 & 47.04 \\

\midrule
        \textit{\textbf{Anthropic}} \\
Claude-Sonnet-4.5                    & 51.24 & 24.56 & 34.04 \\
 \rowcolor{softpeach}
Claude-Sonnet-4.5$_\texttt{CoT}$     & 75.88 & 44.40 & 54.89 \\
Claude-Opus-4.5                      & 61.97 & 10.17 & 16.00 \\
 \rowcolor{softpeach}
Claude-Opus-4.5$_\texttt{CoT}$       & 76.68 & \textbf{44.50} & \textbf{55.44} \\
Claude-Opus-4.1                      & 54.40 & 19.31 & 28.50 \\
 \rowcolor{softpeach}
Claude-Opus-4.1$_\texttt{CoT}$       & \textbf{76.95} & 39.75 & 49.81 \\
Claude-Sonnet-4                      & 51.40 & 21.92 & 31.80 \\
 \rowcolor{softpeach}
Claude-Sonnet-4$_\texttt{CoT}$       & 74.15 & 39.09 & 49.40 \\
Claude-Opus-4                        & 52.02 & 22.45 & 32.16 \\
 \rowcolor{softpeach}
Claude-Opus-4$_\texttt{CoT}$         & 73.10 & 40.44 & 50.18 \\

\midrule
        \textit{\textbf{Google}} \\
Gemini-3-Flash-Preview               & 67.22 & 21.30 & 31.34 \\
 \rowcolor{softpeach}
Gemini-3-Flash-Preview$_\texttt{CoT}$ & 75.90 & 40.54 & 50.19 \\
Gemini-2.5-Flash                     & 62.77 & 18.20 & 25.10 \\
 \rowcolor{softpeach}
Gemini-2.5-Flash$_\texttt{CoT}$      & 54.60 & 27.71 & 36.55 \\
Gemini-2.5-Flash-Lite                & 32.68 & 9.31 & 14.04 \\
 \rowcolor{softpeach}
Gemini-2.5-Flash-Lite$_\texttt{CoT}$ & 32.26 & 22.54 & 29.24 \\

    \bottomrule
    \end{tabular}}
    \caption{Expanded proprietary LLMs results on OmanicBench.}
    \label{tab:main_app_result}
\end{table}

\begin{table}[t]
    \centering
    \resizebox{\columnwidth}{!}{
    \begin{tabular}{lccc}
    \toprule
         & \textbf{MCQ} & \textbf{Exact Match} & \textbf{F1-Score} \\
    \midrule
    \textit{\textbf{Meta}} \\
    LLaMA-3.3-70B                        & 40.04 & 11.77 & 20.47 \\
    \rowcolor{softpeach}
    LLaMA-3.3-70B$_\texttt{CoT}$         & 59.57 & 31.33 & 39.43 \\
    LLaMA-3-8B                           & 25.23 & 4.96  & 8.78 \\
    \rowcolor{softpeach}
    LLaMA-3-8B$_\texttt{CoT}$            & 42.50 & 16.55 & 22.78 \\
    LLaMA-3-70B                          & 38.47 & 12.10 & 18.84 \\
    \rowcolor{softpeach}
    LLaMA-3-70B$_\texttt{CoT}$           & 57.39 & 29.09 & 37.49 \\

    \midrule
    \textit{\textbf{Alibaba}} \\
    Qwen3-32B                            & 52.22 & 12.10 & 18.35 \\
    \rowcolor{softpeach}
    Qwen3-32B$_\texttt{CoT}$             & 54.86 & 30.37 & 38.98 \\
    Qwen3-8B                             & 25.65 & 9.26  & 13.77 \\
    \rowcolor{softpeach}
    Qwen3-8B$_\texttt{CoT}$              & 56.44 & 21.76 & 29.97 \\
    Qwen2.5-72B                          & 42.19 & 13.24 & 19.43 \\
    \rowcolor{softpeach}
    Qwen2.5-72B$_\texttt{CoT}$           & 60.81 & 32.06 & 41.31 \\
    Qwen2.5-7B                           & 30.51 & 7.76  & 14.31 \\
    \rowcolor{softpeach}
    Qwen2.5-7B$_\texttt{CoT}$            & 46.85 & 19.44 & 25.60 \\

    \midrule
    \textit{\textbf{DeepSeek}} \\
    DeepSeek-V3.2                        & 43.33 & 12.82 & 19.36 \\
    \rowcolor{softpeach}
    DeepSeek-V3.2$_\texttt{CoT}$         & 70.94 & 35.92 & 46.19 \\
    DeepSeek-R1-Distill-LLaMA-70B        & \textbf{75.92} & 1.72  & 13.71 \\
    \rowcolor{softpeach}
    DeepSeek-R1-Distill-LLaMA-70B$_\texttt{CoT}$ & 72.67 & \textbf{41.27} & \textbf{51.08} \\
    DeepSeek-R1-Distill-Qwen-32B         & 69.52 & 11.92 & 22.99 \\
    \rowcolor{softpeach}
    DeepSeek-R1-Distill-Qwen-32B$_\texttt{CoT}$ & 69.29 & 35.51 & 46.00 \\
    \bottomrule
    \end{tabular}}
    \caption{Expanded Open-source LLMs results on OmanicBench.}
    \label{tab:main_app2_result}
\end{table}

\begin{table}[t]
    \centering
    \scriptsize
    \resizebox{\columnwidth}{!}{
    \begin{tabular}{lrrrr}
    \toprule
    \textbf{Model} & \textbf{Step 1} & \textbf{Step 2} & \textbf{Step 3} & \textbf{Step 4} \\
    \midrule
    GPT-4o & 89.76 & 66.39 & 72.08 & \textcolor{red}{49.02} \\
    GPT-5.1 & 87.49 & 70.63 & 72.49 & \textcolor{red}{51.71} \\
    GPT-5.2 & 82.01 & 71.04 & 73.73 & \textcolor{red}{53.67} \\
    \midrule
    Claude-Sonnet-4 & 89.97 & 89.04 & 86.35 & \textcolor{red}{67.32} \\
    Claude-Opus-4 & 94.62 & 91.93 & 90.49 & \textcolor{red}{72.80} \\
    Claude-Opus-4.1 & 95.35 & 92.55 & 90.69 & \textcolor{red}{72.08} \\
    Claude-Sonnet-4.5 & 91.52 & 88.21 & 86.45 & \textcolor{red}{70.53} \\
    Claude-Opus-4.5 & 93.28 & 91.73 & 88.73 & \textcolor{red}{74.56} \\
    \midrule
    Gemini-2.5-Flash-Lite & 72.80 & 66.29 & 68.77 & \textcolor{red}{45.50} \\
    Gemini-2.5-Flash & 91.52 & 94.52 & 90.80 & \textcolor{red}{75.59} \\
    Gemini-3-Flash-Preview & 97.31 & 96.69 & 92.76 & \textcolor{red}{79.42} \\
    \midrule
    Qwen3-Max & 87.69 & 85.42 & 85.94 & \textcolor{red}{67.11} \\
    \bottomrule
    \end{tabular}}
    \caption{Step-level accuracy under direct prompting on OmanicBench.}
    \label{tab:step-level-direct}
\end{table}

\begin{table}[t]
    \centering
    \scriptsize
    \resizebox{\columnwidth}{!}{
    \begin{tabular}{lrrrr}
    \toprule
    \textbf{Model} & \textbf{Step 1} & \textbf{Step 2} & \textbf{Step 3} & \textbf{Step 4} \\
    \midrule
    GPT-4o & 92.55 & 94.11 & 89.35 & \textcolor{red}{75.49} \\
    GPT-5.1 & 94.83 & 94.73 & 91.93 & \textcolor{red}{78.70} \\
    GPT-5.2 & 90.69 & 94.11 & 90.18 & \textcolor{red}{77.77} \\
    \midrule
    Claude-Sonnet-4 & 93.38 & 96.79 & 92.45 & \textcolor{red}{80.77} \\
    Claude-Opus-4 & 96.07 & 95.66 & 92.66 & \textcolor{red}{79.94} \\
    Claude-Opus-4.1 & 96.79 & 95.97 & 92.86 & \textcolor{red}{80.97} \\
    Claude-Sonnet-4.5 & 95.86 & 96.48 & 93.17 & \textcolor{red}{81.80} \\
    Claude-Opus-4.5 & 95.76 & 96.59 & 93.17 & \textcolor{red}{80.35} \\
    \midrule
    Gemini-2.5-Flash-Lite & 72.70 & 47.36 & 40.95 & \textcolor{red}{26.78} \\
    Gemini-2.5-Flash & 81.39 & 82.63 & 73.11 & \textcolor{red}{56.26} \\
    Gemini-3-Flash-Preview & 96.17 & 95.76 & 92.45 & \textcolor{red}{80.87} \\
    \midrule
    Qwen3-Max & 91.93 & 93.17 & 91.21 & \textcolor{red}{78.08} \\
    \bottomrule
    \end{tabular}}
    \caption{Step-level accuracy under CoT prompting on OmanicBench.}
    \label{tab:step-level-cot}
\end{table}


\clearpage
\subsection{Full results on Knowledge Floor and the Error Propagation}
For the step-wise error rates under chain evaluation, we exclude single-hop questions that can be answered without relying on preceding steps, since they do not reflect dependency-sensitive error propagation. For example, under the Bridge topology, the third hop is omitted because it can be answered independently of earlier hops.

\begin{figure}[h]
    \centering
    \includegraphics[width=\columnwidth]{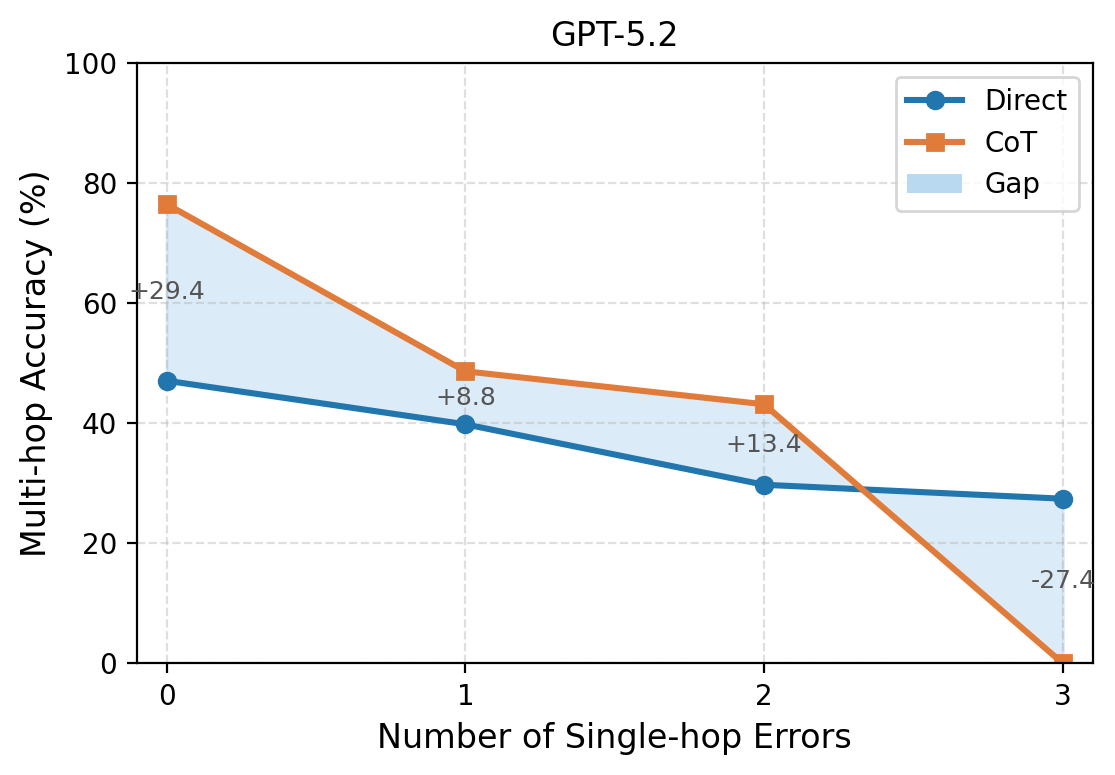}
    \caption{Multi-hop accuracy by number of single-hop errors for GPT-5.2.}
\end{figure}

\begin{figure}[h]
    \centering
    \includegraphics[width=\columnwidth]{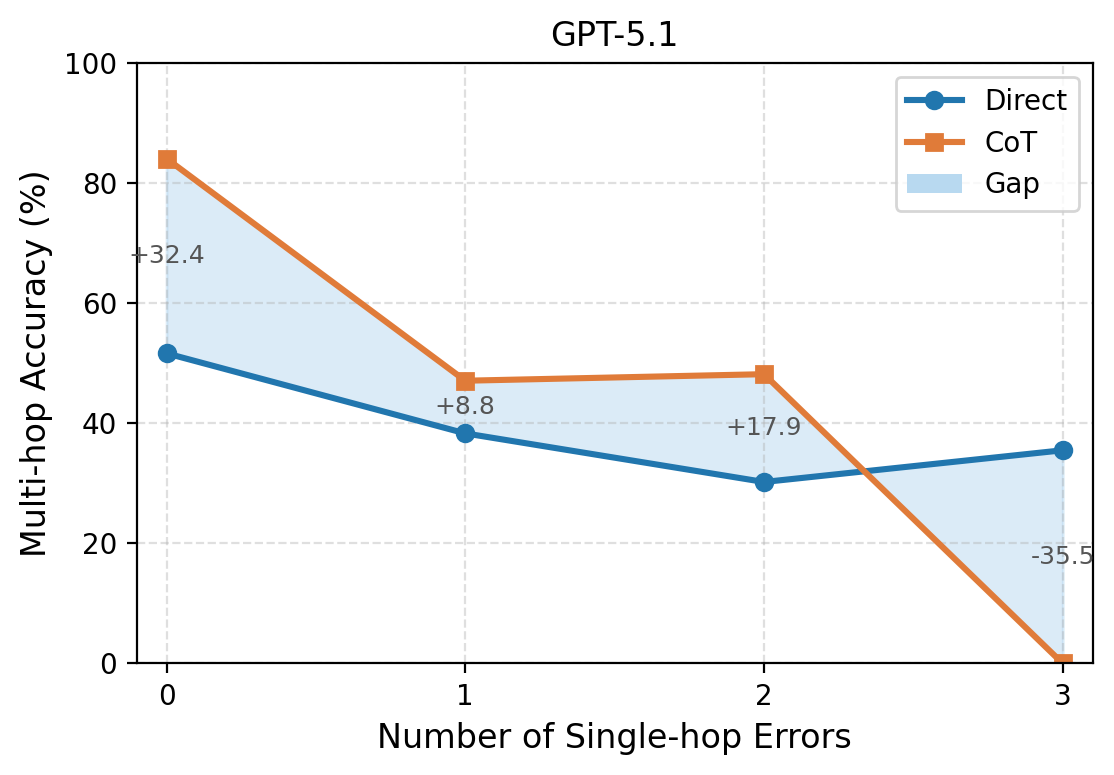}
    \caption{Multi-hop accuracy by number of single-hop errors for GPT-5.1.}
\end{figure}

\begin{figure}[h]
    \centering
    \includegraphics[width=\columnwidth]{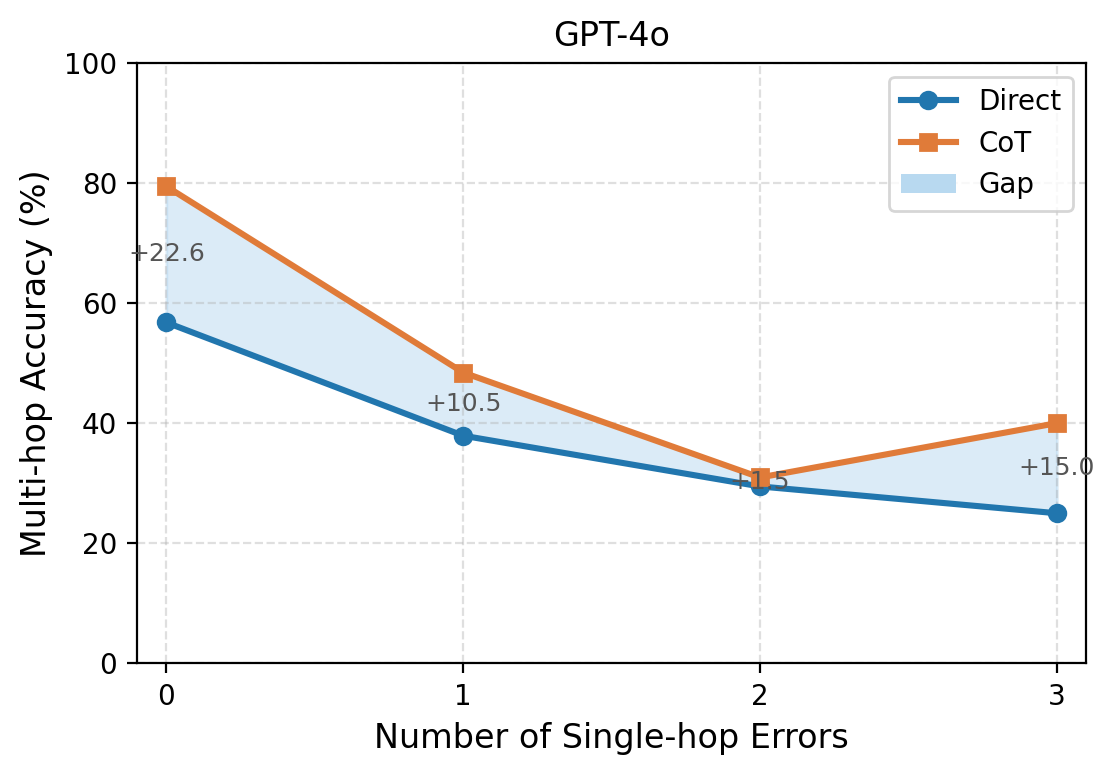}
    \caption{Multi-hop accuracy by number of single-hop errors for GPT-4o.}
\end{figure}

\begin{figure}[h]
    \centering
    \includegraphics[width=\columnwidth]{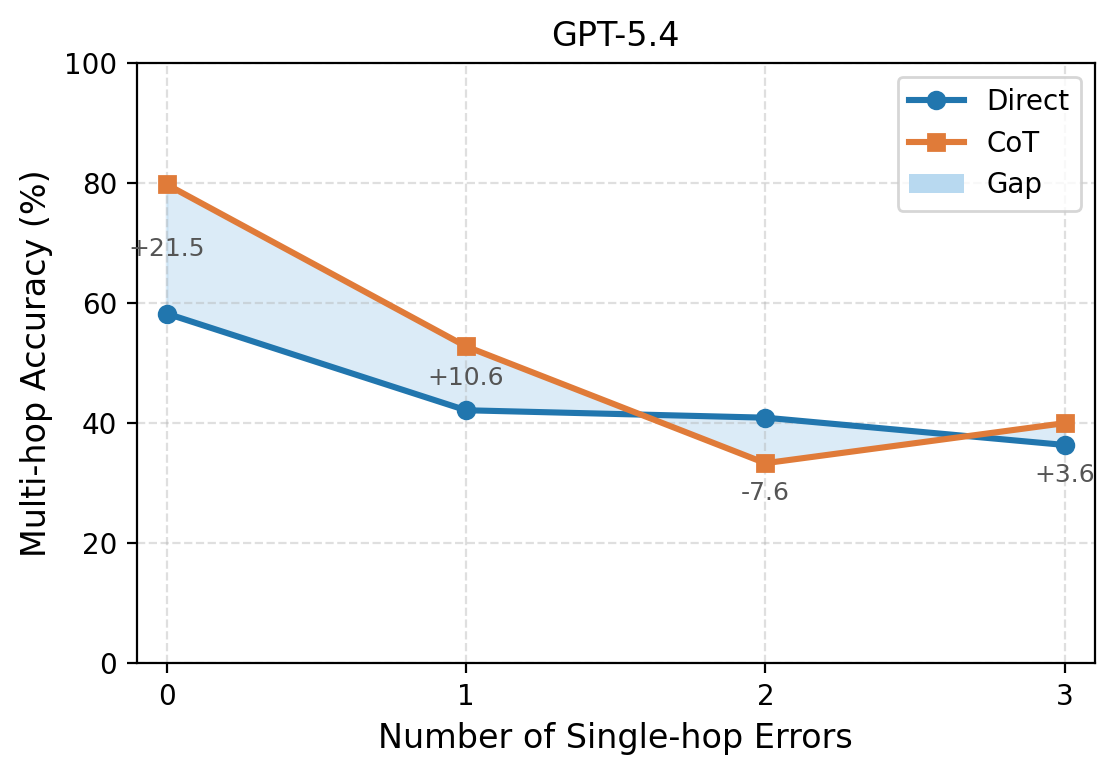}
    \caption{Multi-hop accuracy by number of single-hop errors for GPT-5.4.}
\end{figure}

\begin{figure}[h]
    \centering
    \includegraphics[width=\columnwidth]{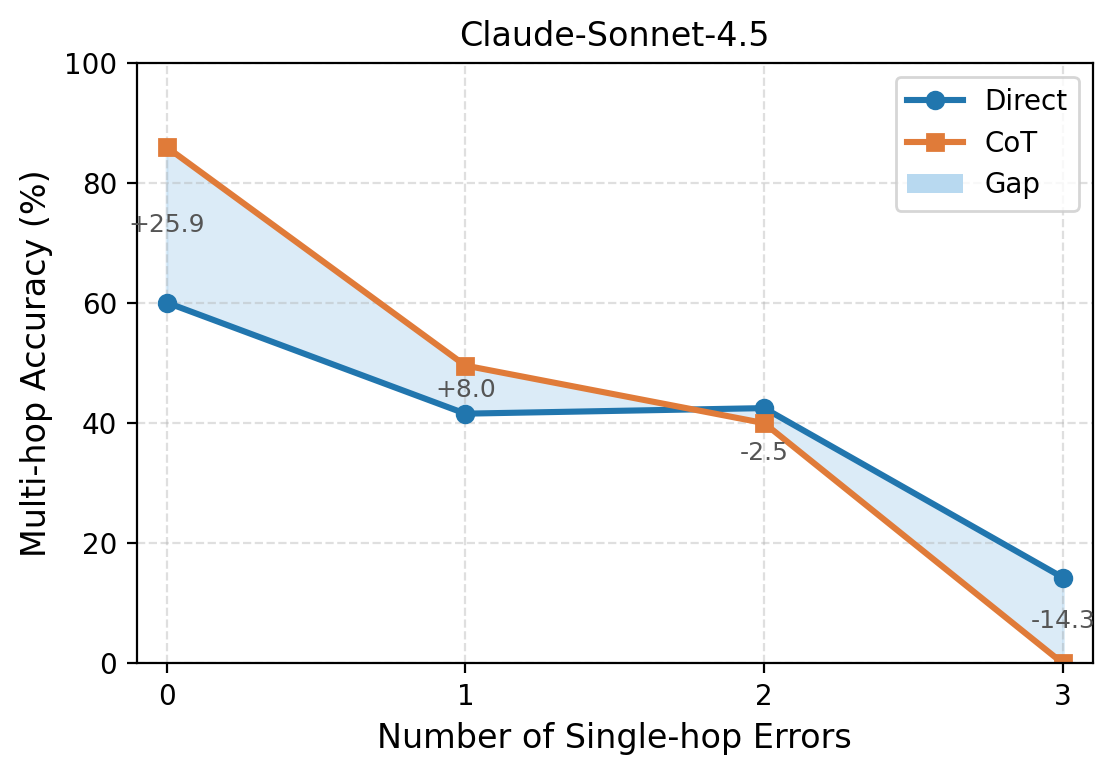}
    \caption{Multi-hop accuracy by number of single-hop errors for Claude-Sonnet-4.5.}
\end{figure}

\begin{figure}[h]
    \centering
    \includegraphics[width=\columnwidth]{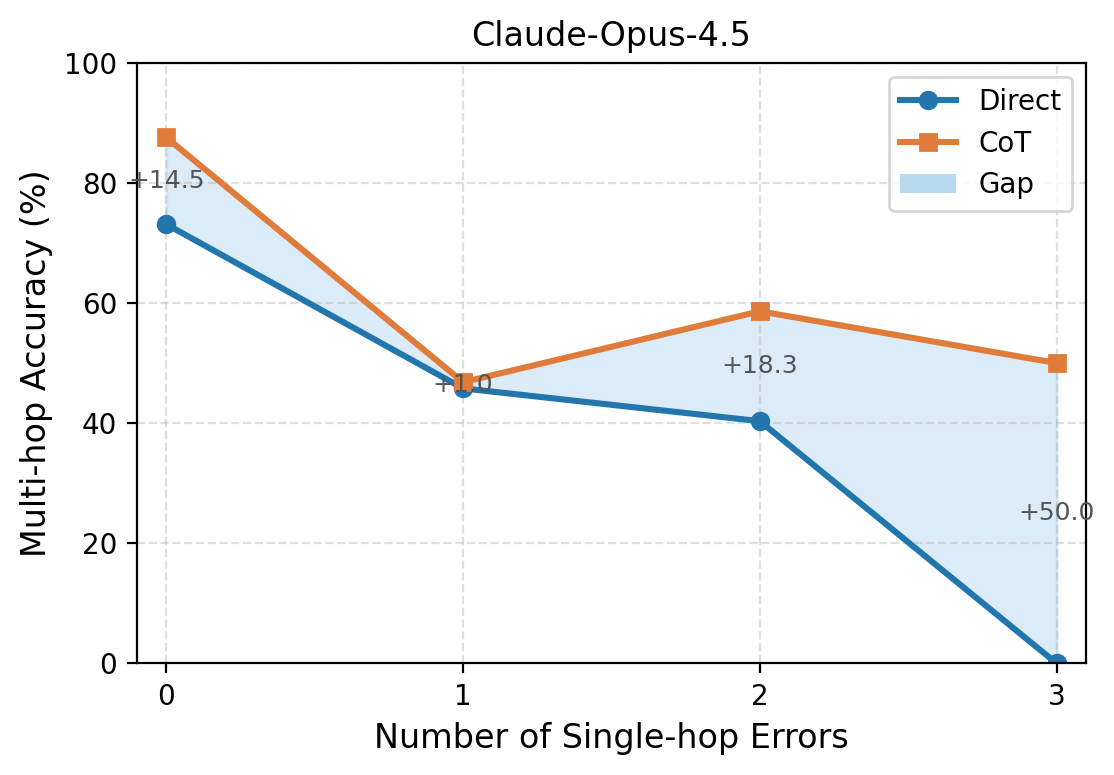}
    \caption{Multi-hop accuracy by number of single-hop errors for Claude-Opus-4.5.}
\end{figure}

\begin{figure}[h]
    \centering
    \includegraphics[width=\columnwidth]{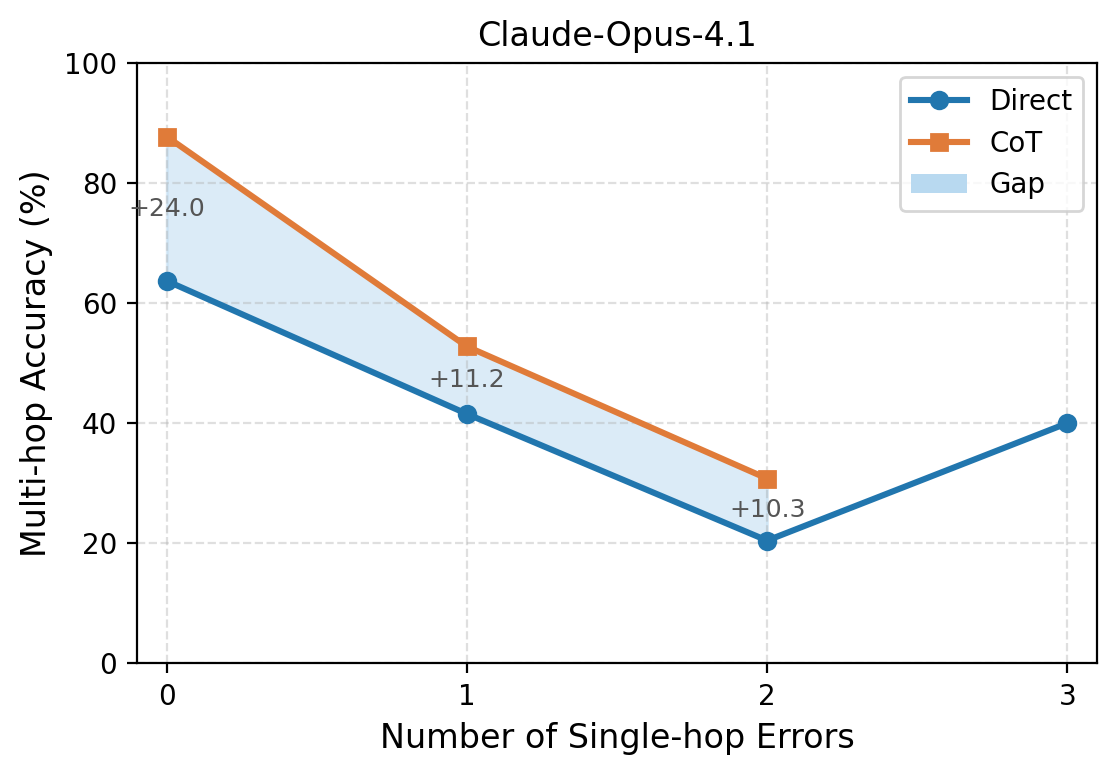}
    \caption{Multi-hop accuracy by number of single-hop errors for Claude-Opus-4.1.}
\end{figure}

\begin{figure}[h]
    \centering
    \includegraphics[width=\columnwidth]{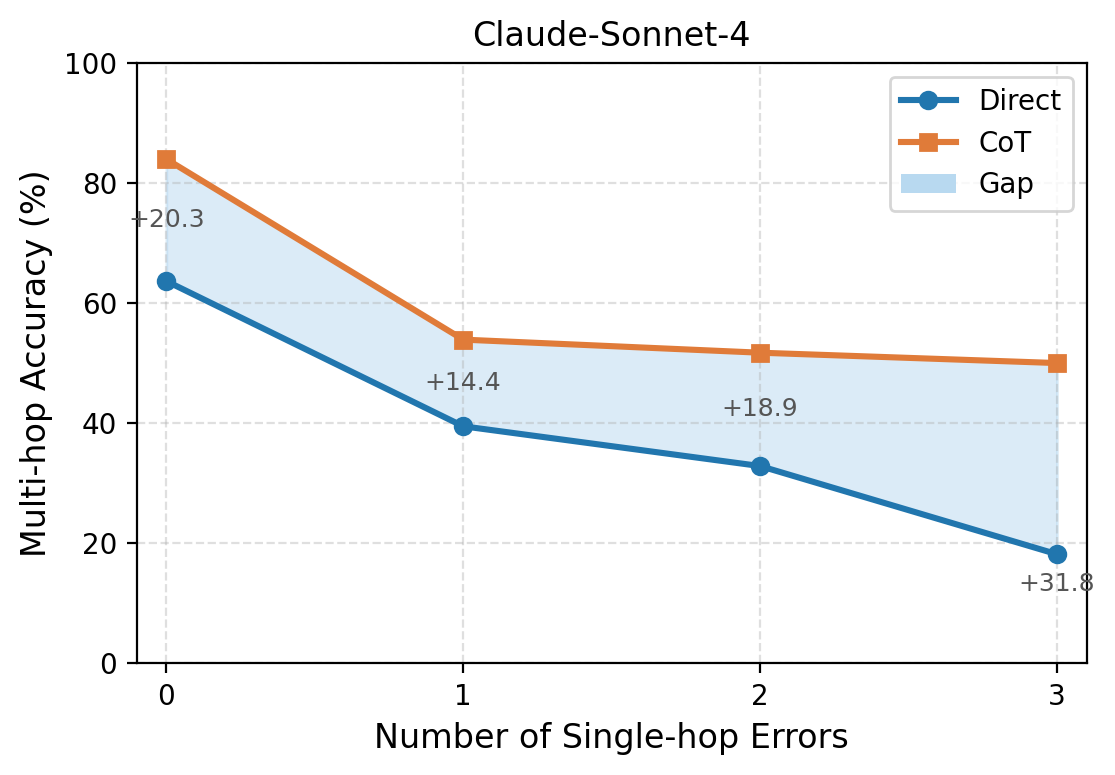}
    \caption{Multi-hop accuracy by number of single-hop errors for Claude-Sonnet-4.}
\end{figure}

\begin{figure}[h]
    \centering
    \includegraphics[width=\columnwidth]{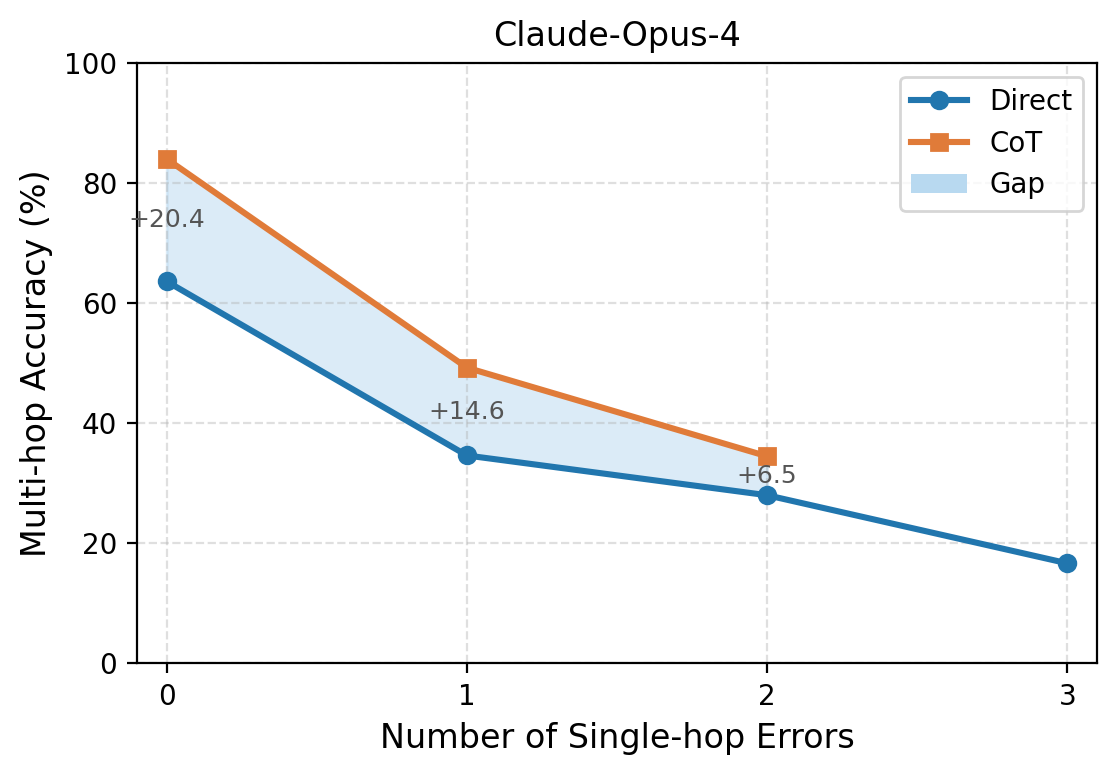}
    \caption{Multi-hop accuracy by number of single-hop errors for Claude-Opus-4.}
\end{figure}

\begin{figure}[h]
    \centering
    \includegraphics[width=\columnwidth]{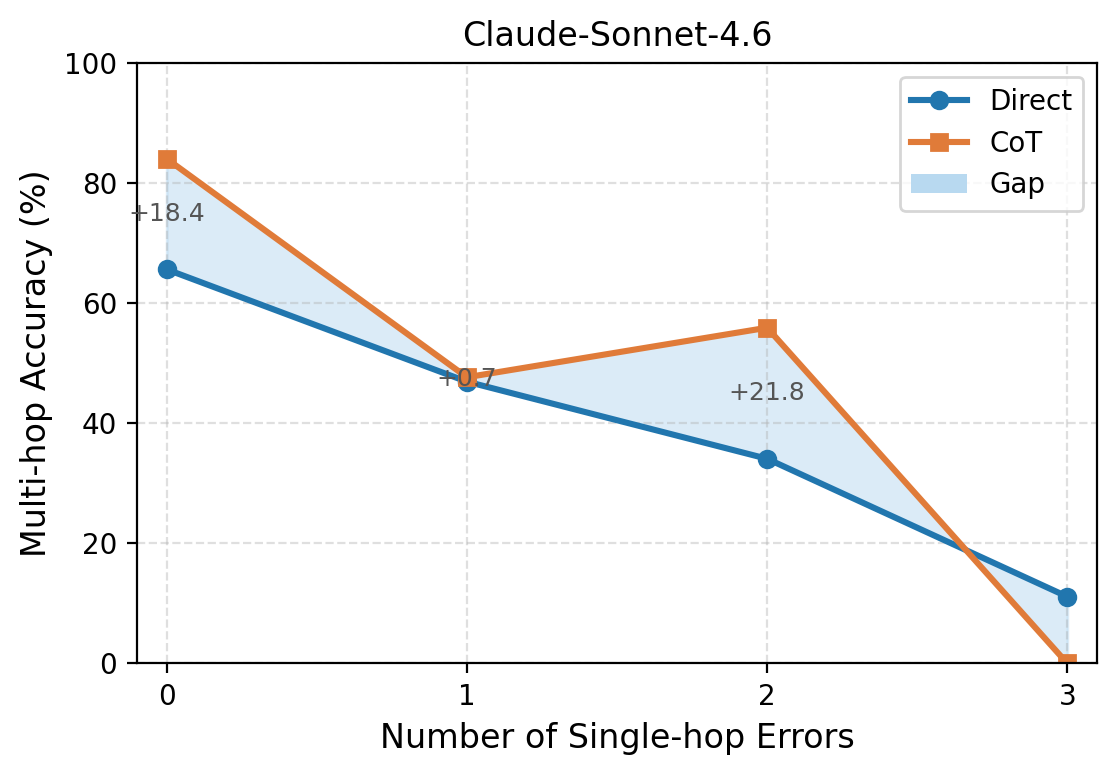}
    \caption{Multi-hop accuracy by number of single-hop errors for Claude-Sonnet-4.6.}
\end{figure}

\begin{figure}[h]
    \centering
    \includegraphics[width=\columnwidth]{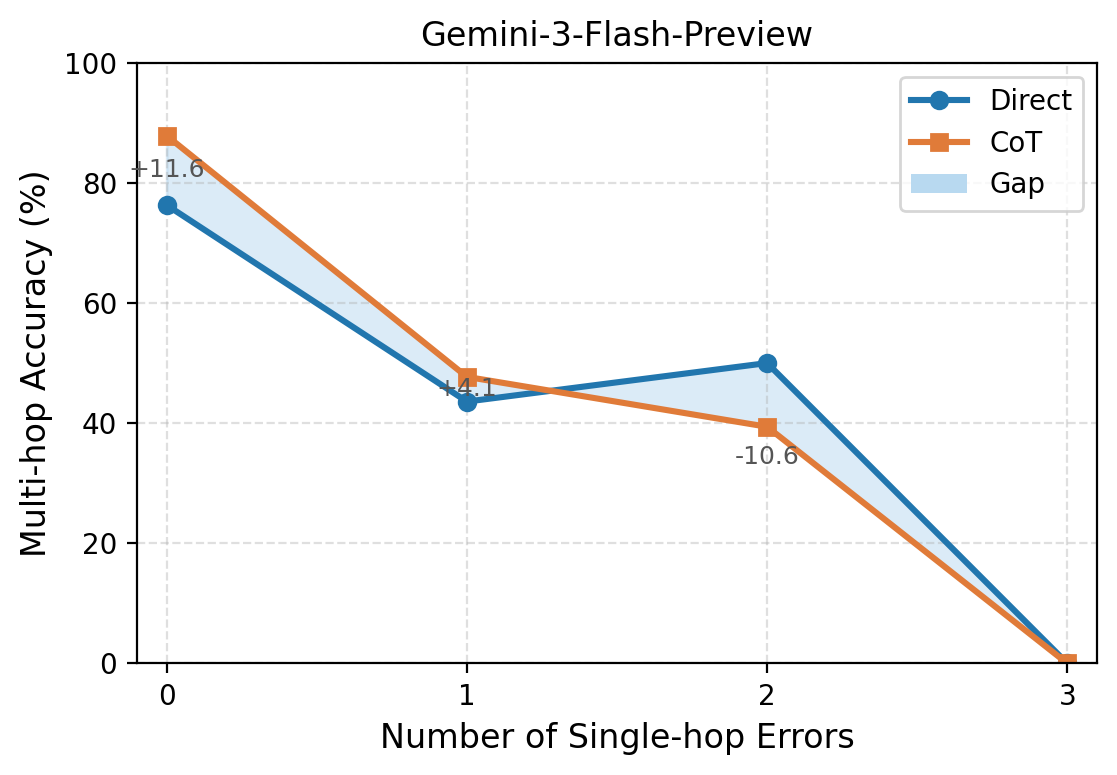}
    \caption{Multi-hop accuracy by number of single-hop errors for Gemini-3-Flash-Preview.}
\end{figure}

\begin{figure}[h]
    \centering
    \includegraphics[width=\columnwidth]{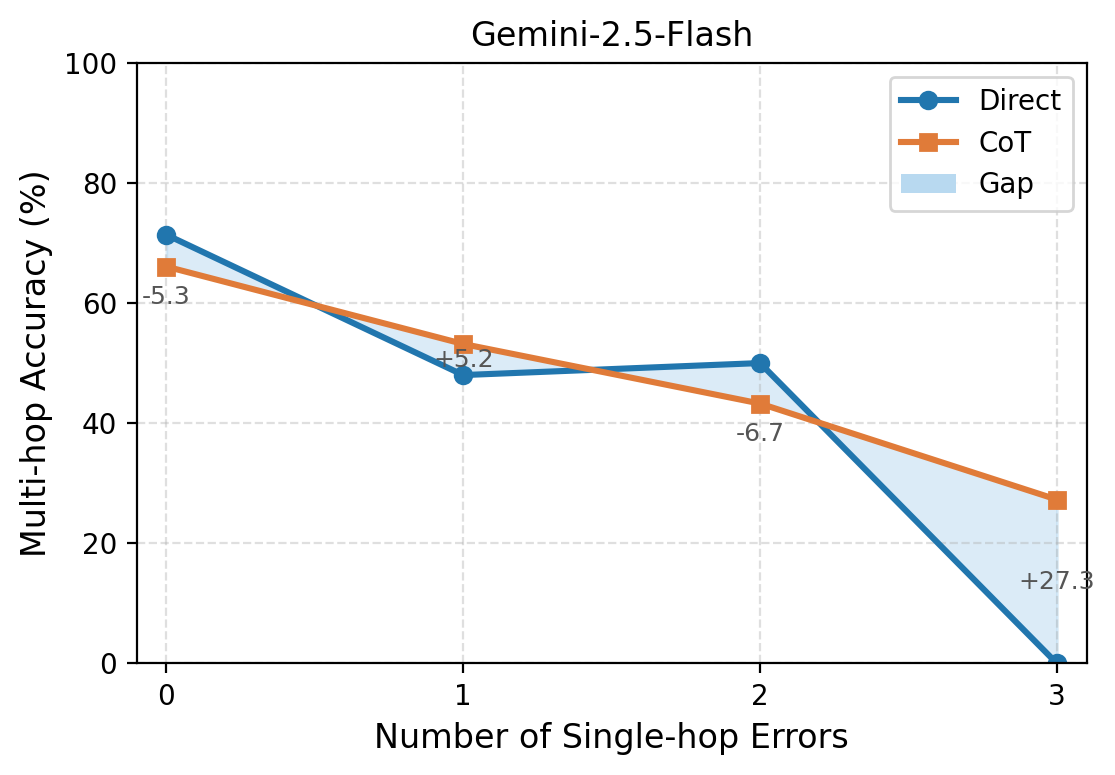}
    \caption{Multi-hop accuracy by number of single-hop errors for Gemini-2.5-Flash.}
\end{figure}

\begin{figure}[h]
    \centering
    \includegraphics[width=\columnwidth]{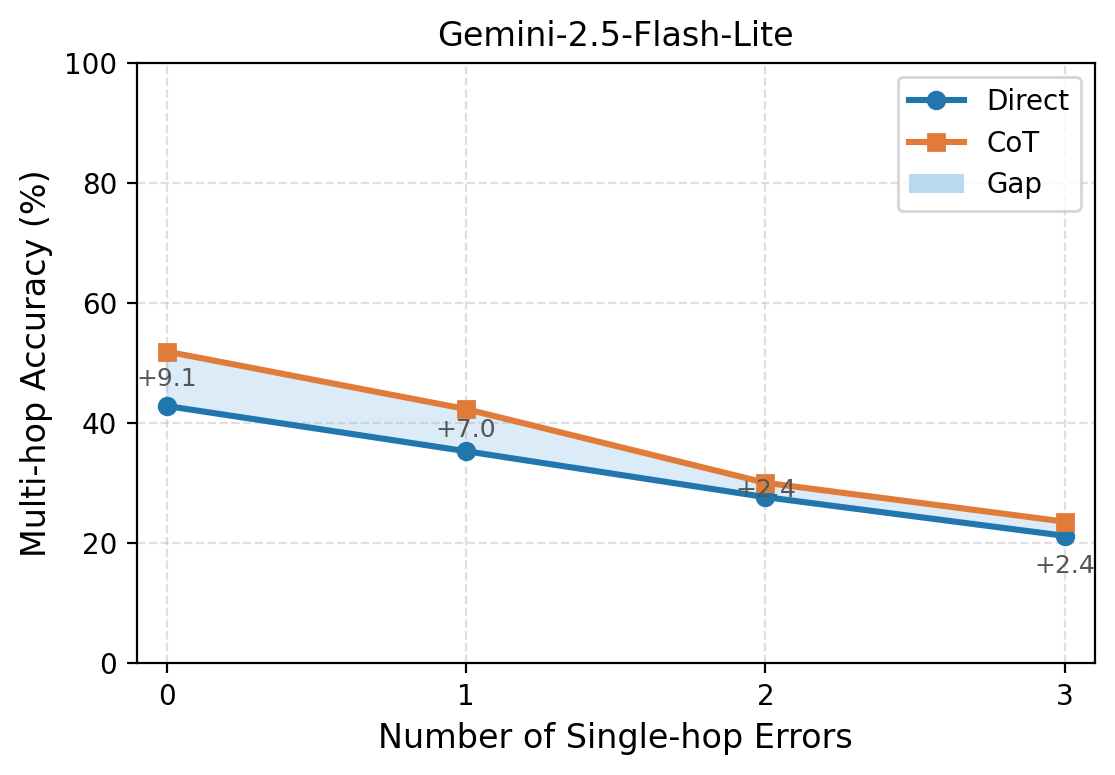}
    \caption{Multi-hop accuracy by number of single-hop errors for Gemini-2.5-Flash-Lite.}
\end{figure}

\begin{figure}[h]
    \centering
    \includegraphics[width=\columnwidth]{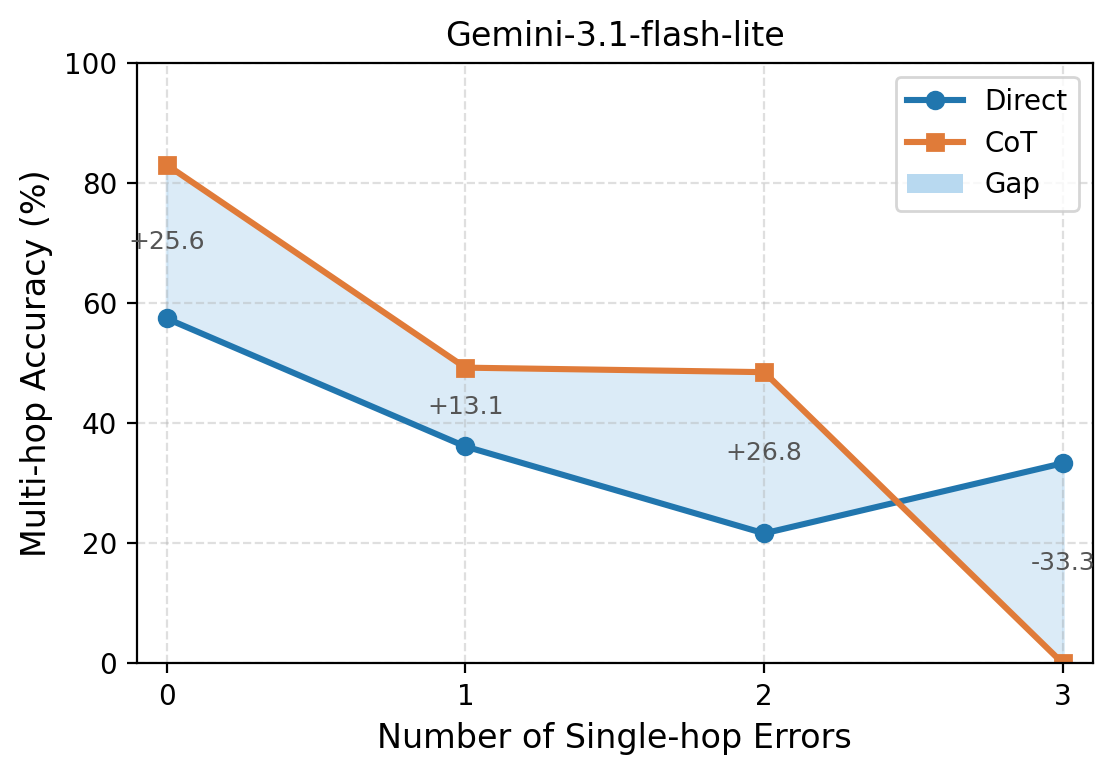}
    \caption{Multi-hop accuracy by number of single-hop errors for Gemini-3.1-flash-lite.}
\end{figure}

\begin{figure}[h]
    \centering
    \includegraphics[width=\columnwidth]{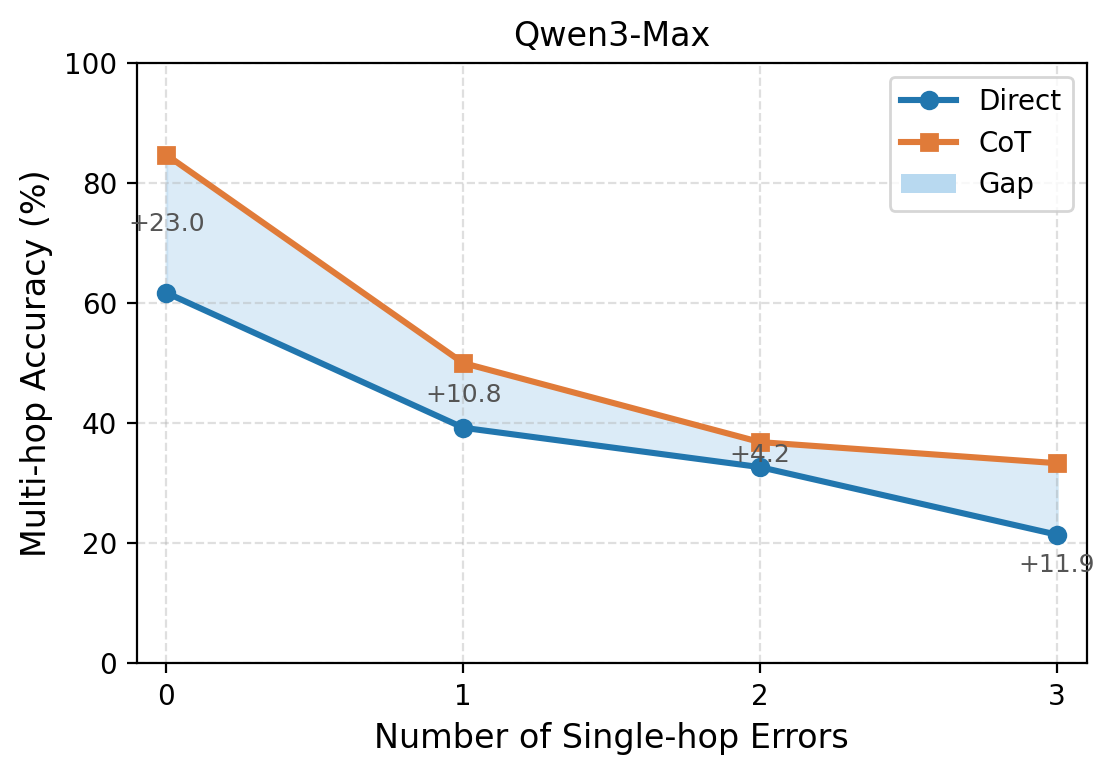}
    \caption{Multi-hop accuracy by number of single-hop errors for Qwen3-Max.}
\end{figure}

\begin{figure}[h]
    \centering
    \includegraphics[width=\columnwidth]{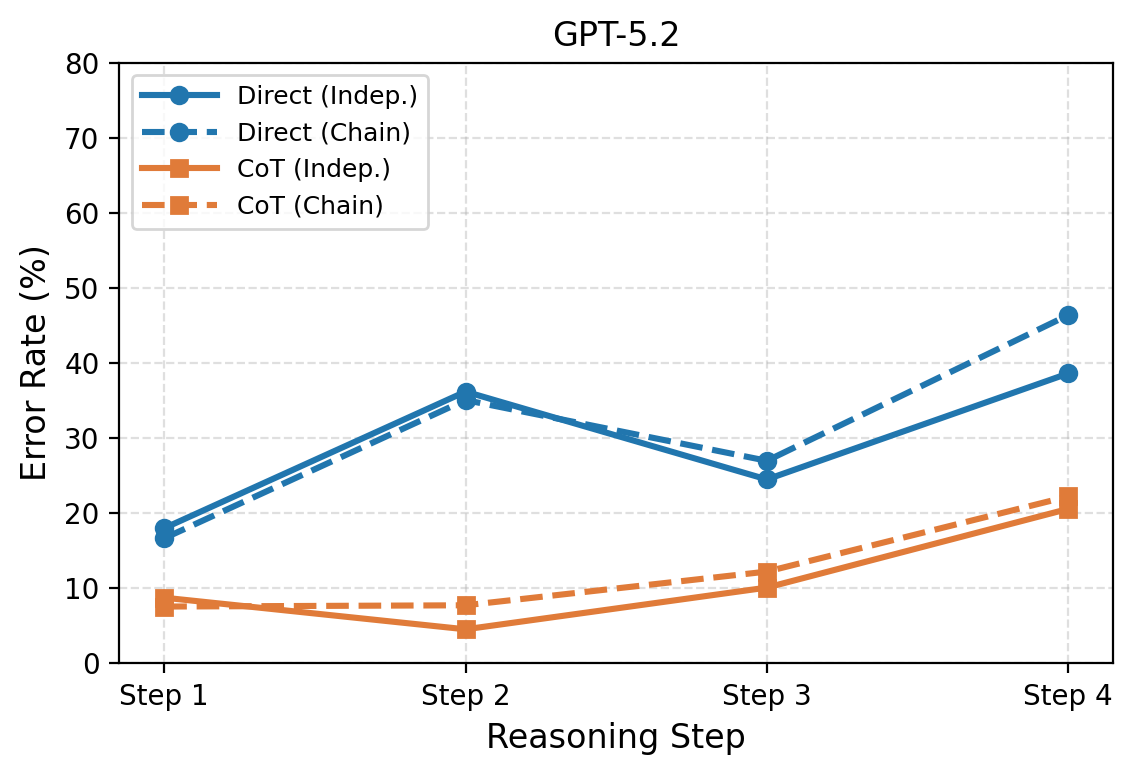}
    \caption{Step-wise error rates under independent and chain evaluation for GPT-5.2.}
\end{figure}

\begin{figure}[h]
    \centering
    \includegraphics[width=\columnwidth]{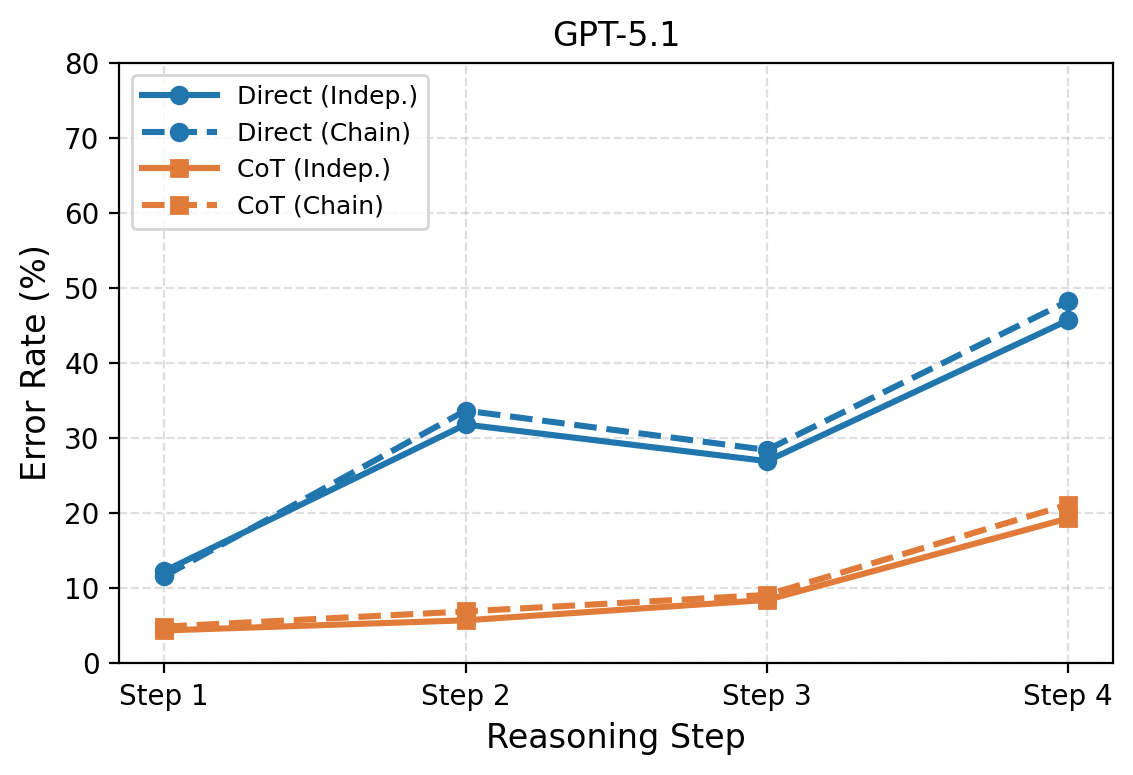}
    \caption{Step-wise error rates under independent and chain evaluation for GPT-5.1.}
\end{figure}

\begin{figure}[h]
    \centering
    \includegraphics[width=\columnwidth]{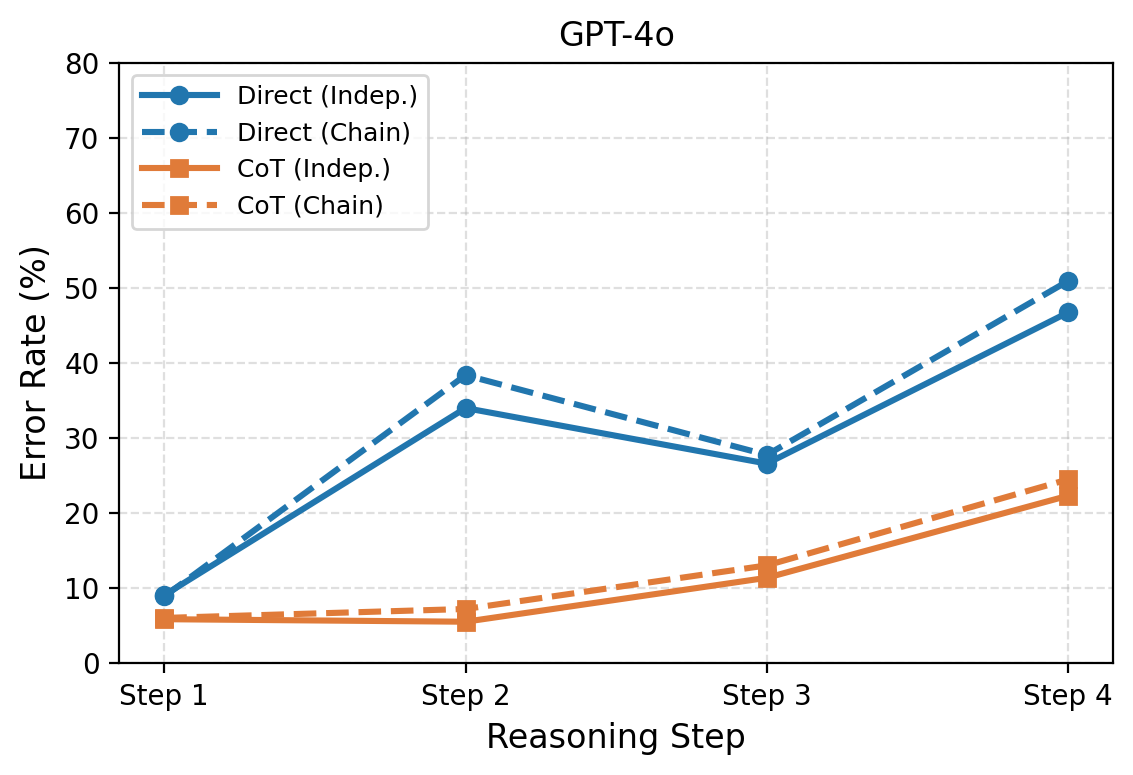}
    \caption{Step-wise error rates under independent and chain evaluation for GPT-4o.}
\end{figure}

\begin{figure}[h]
    \centering
    \includegraphics[width=\columnwidth]{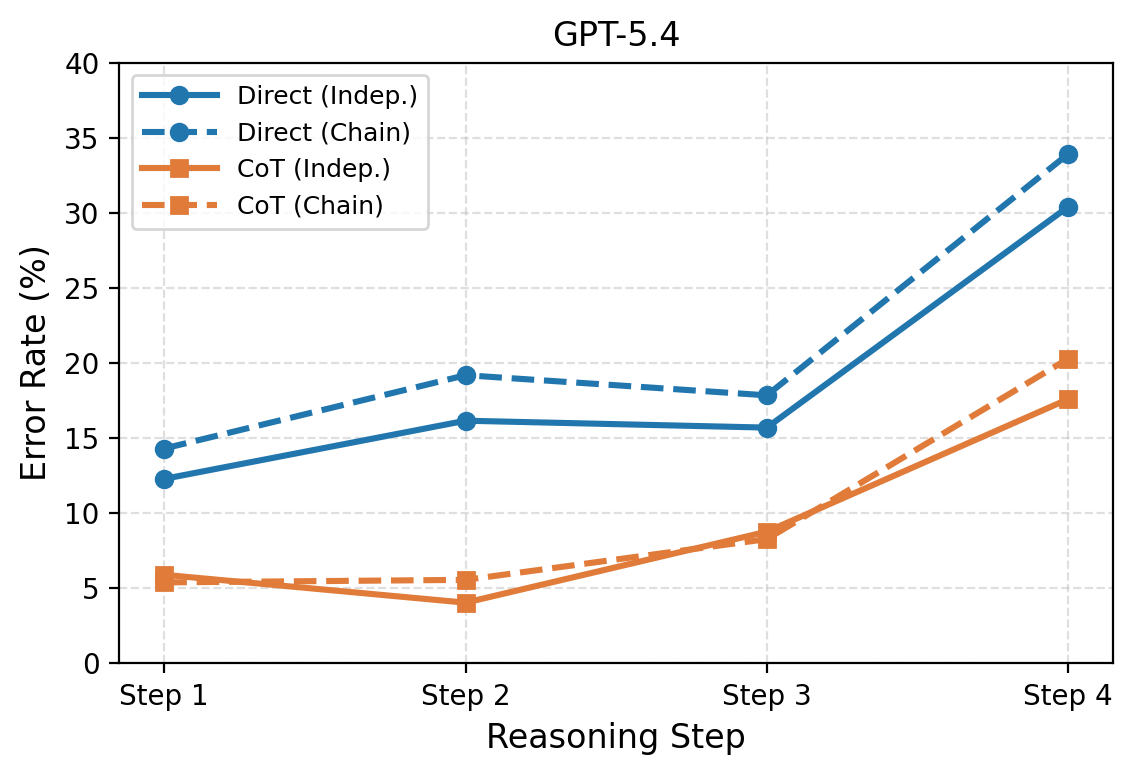}
    \caption{Step-wise error rates under independent and chain evaluation for GPT-5.4.}
\end{figure}

\begin{figure}[h]
    \centering
    \includegraphics[width=\columnwidth]{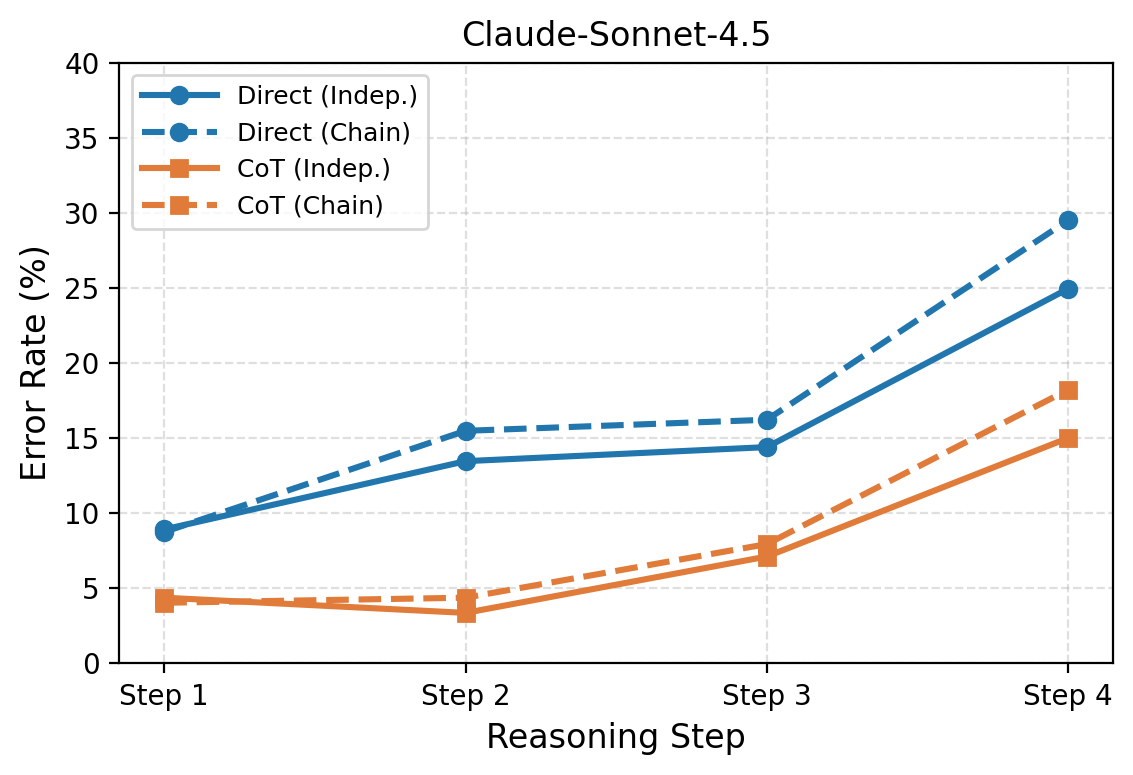}
    \caption{Step-wise error rates under independent and chain evaluation for Claude-Sonnet-4.5.}
\end{figure}

\begin{figure}[h]
    \centering
    \includegraphics[width=\columnwidth]{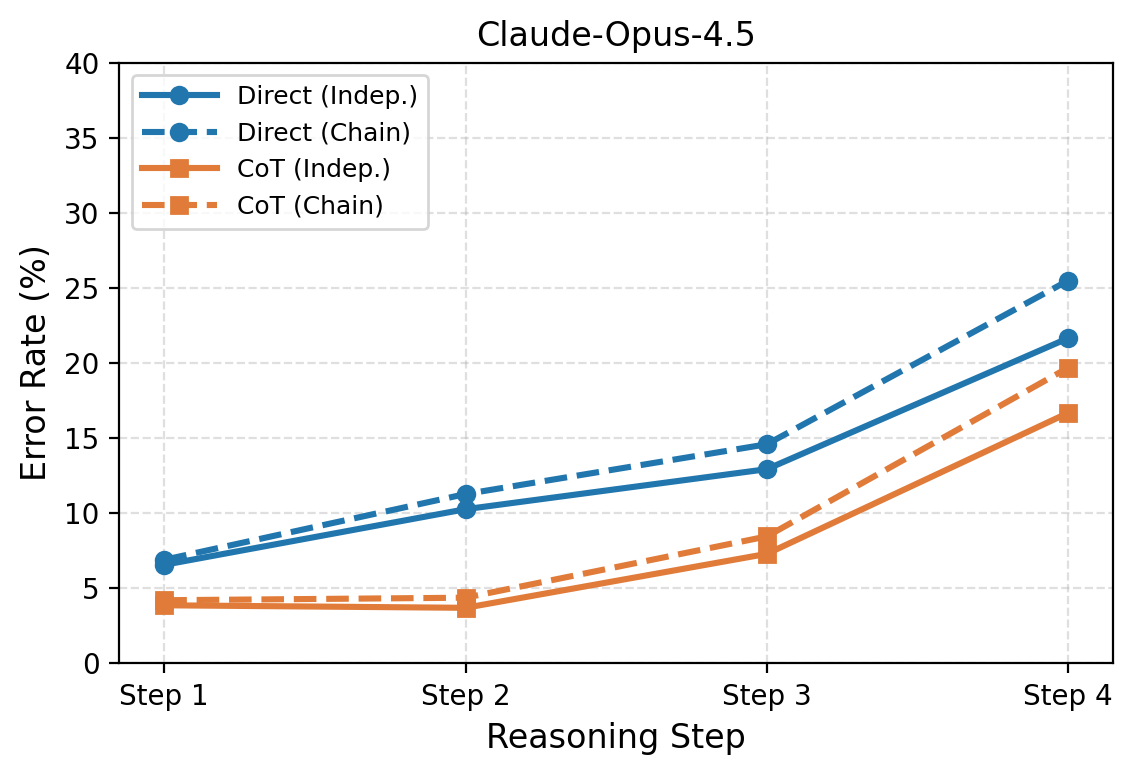}
    \caption{Step-wise error rates under independent and chain evaluation for Claude-Opus-4.5.}
\end{figure}

\begin{figure}[h]
    \centering
    \includegraphics[width=\columnwidth]{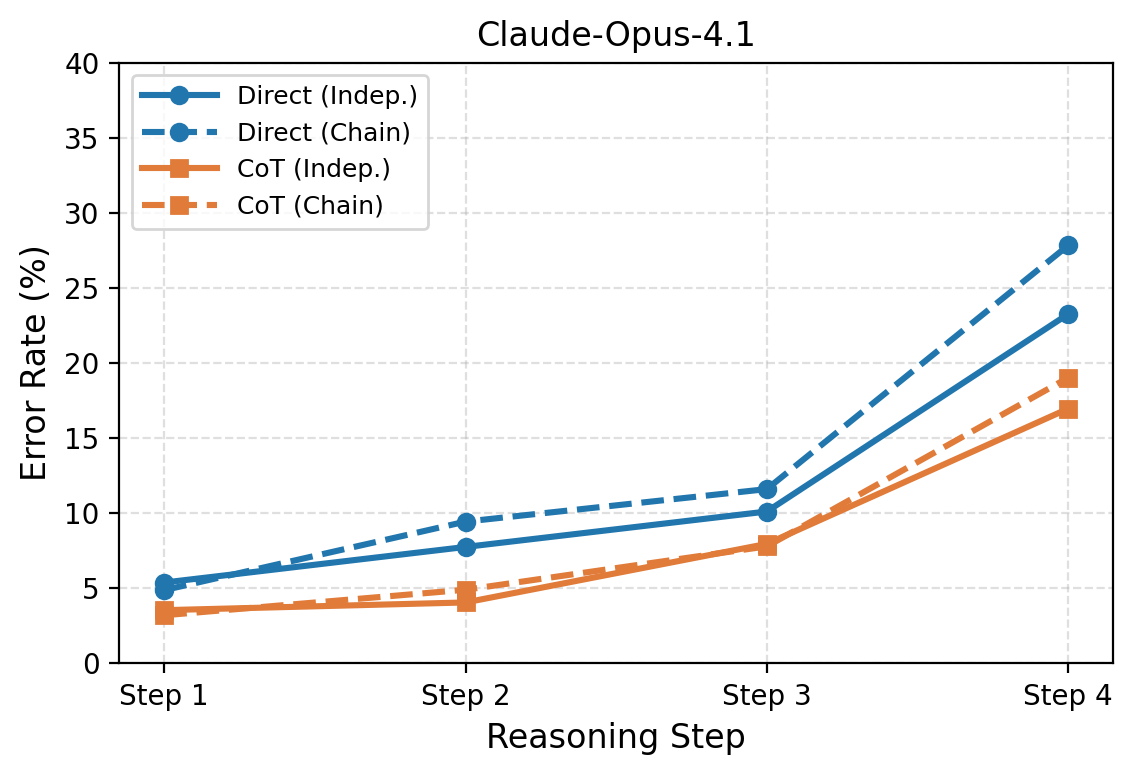}
    \caption{Step-wise error rates under independent and chain evaluation for Claude-Opus-4.1.}
\end{figure}

\begin{figure}[h]
    \centering
    \includegraphics[width=\columnwidth]{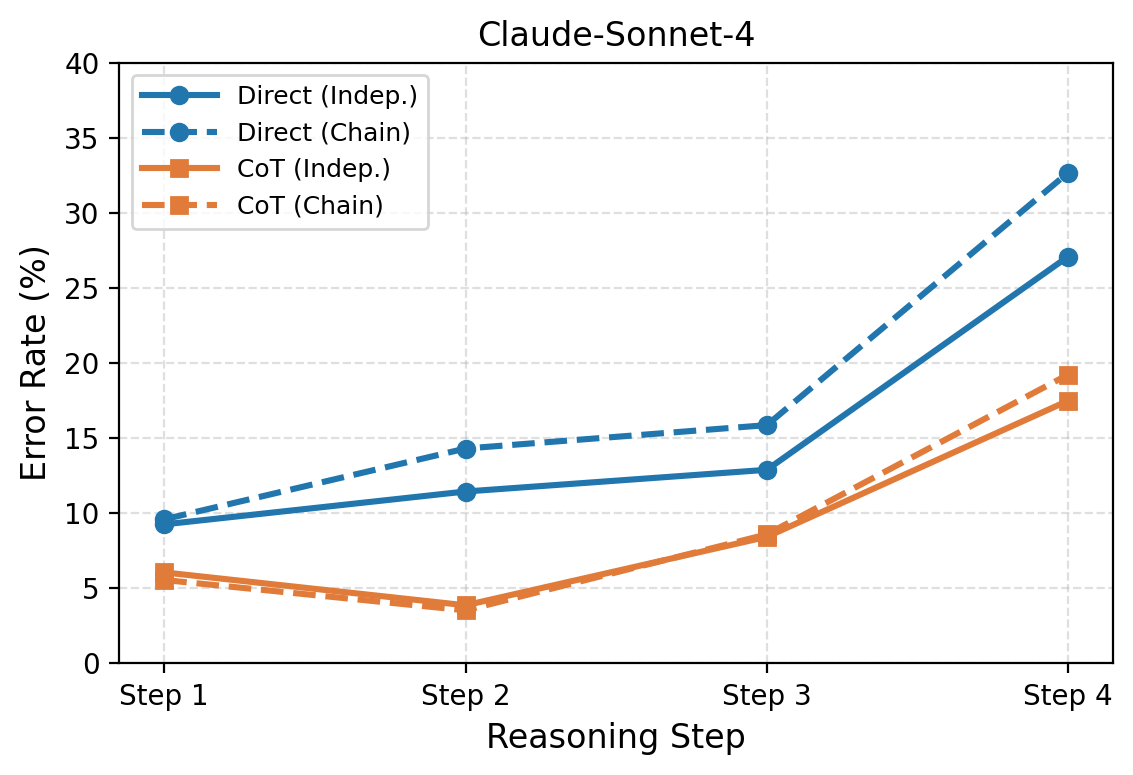}
    \caption{Step-wise error rates under independent and chain evaluation for Claude-Sonnet-4.}
\end{figure}

\begin{figure}[h]
    \centering
    \includegraphics[width=\columnwidth]{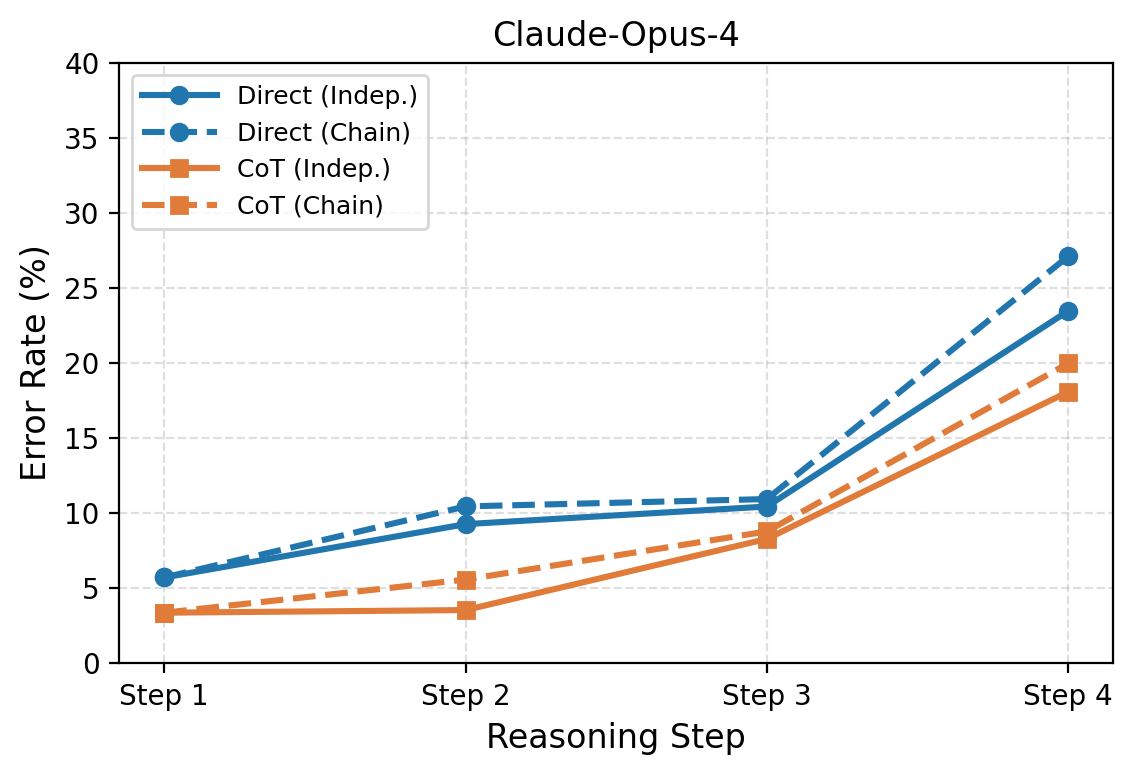}
    \caption{Step-wise error rates under independent and chain evaluation for Claude-Opus-4.}
\end{figure}

\begin{figure}[h]
    \centering
    \includegraphics[width=\columnwidth]{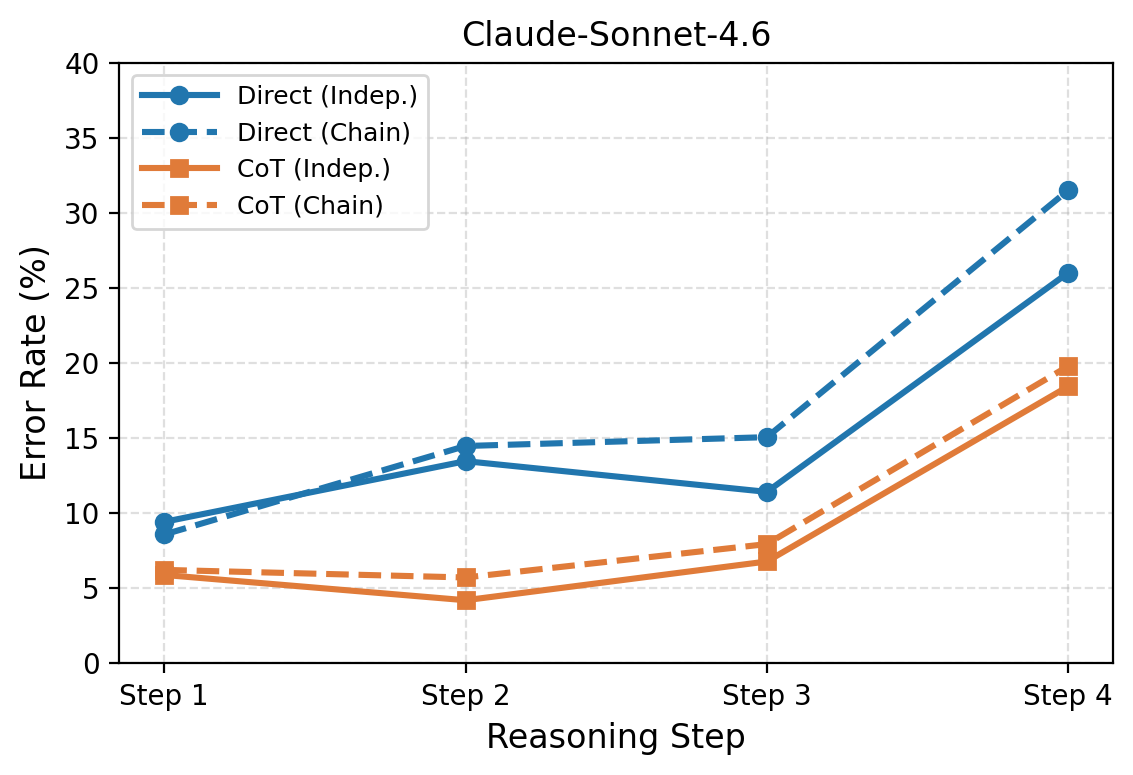}
    \caption{Step-wise error rates under independent and chain evaluation for Claude-Sonnet-4.6.}
\end{figure}

\begin{figure}[h]
    \centering
    \includegraphics[width=\columnwidth]{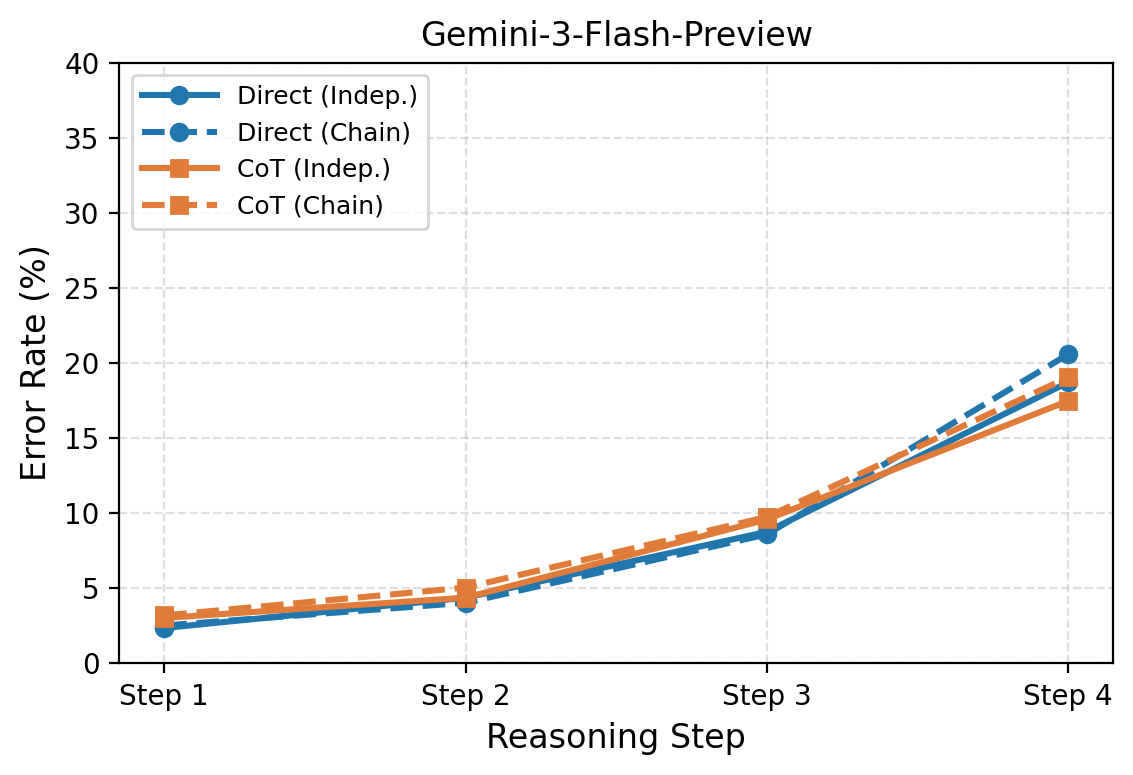}
    \caption{Step-wise error rates under independent and chain evaluation for Gemini-3-Flash-Preview.}
\end{figure}

\begin{figure}[h]
    \centering
    \includegraphics[width=\columnwidth]{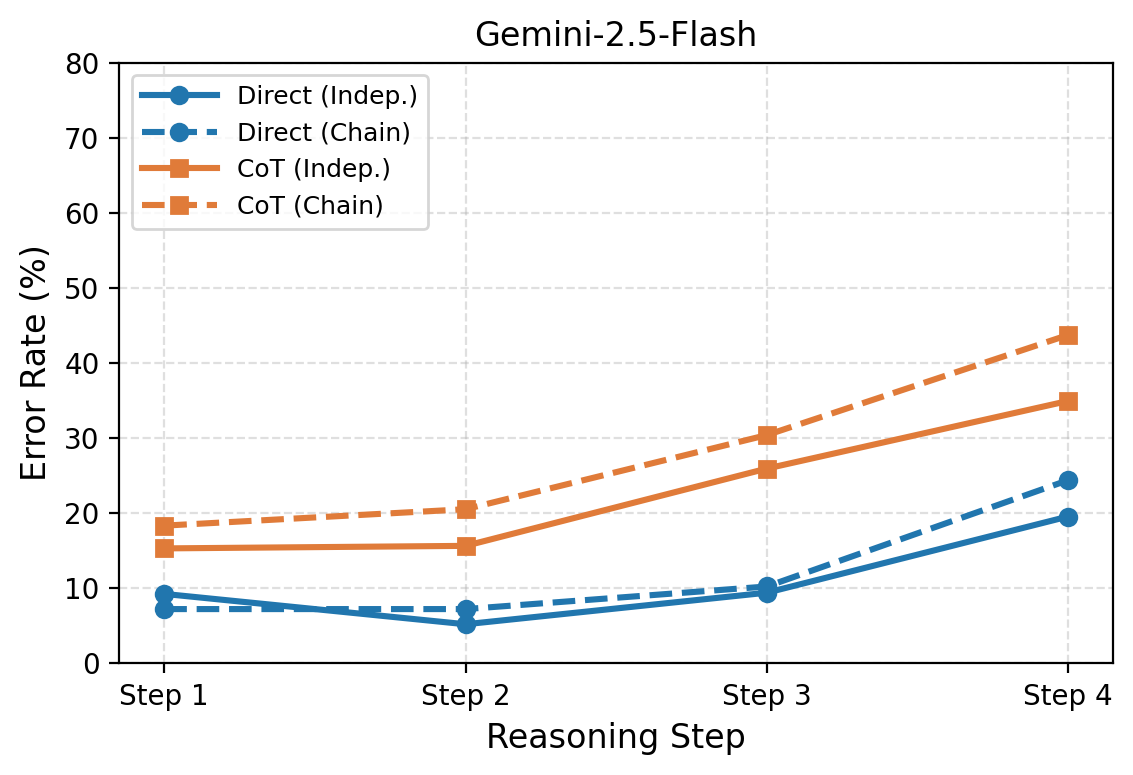}
    \caption{Step-wise error rates under independent and chain evaluation for Gemini-2.5-Flash.}
\end{figure}

\begin{figure}[h]
    \centering
    \includegraphics[width=\columnwidth]{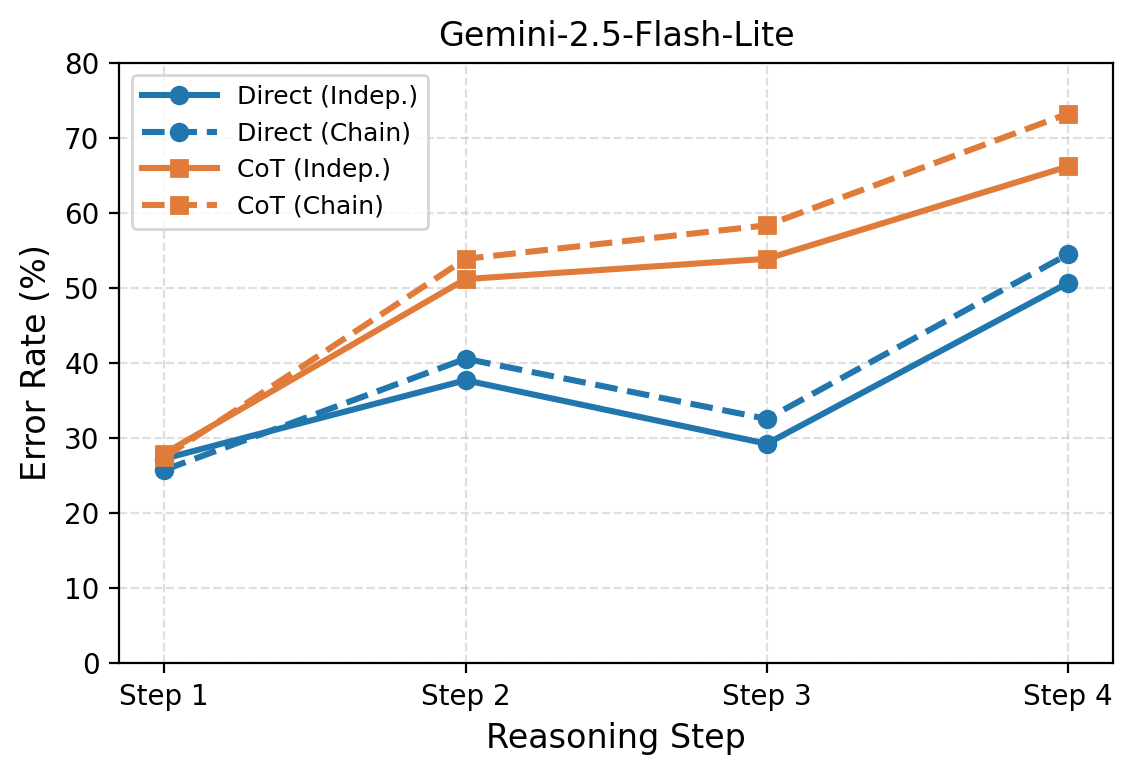}
    \caption{Step-wise error rates under independent and chain evaluation for Gemini-2.5-Flash-Lite.}
\end{figure}

\begin{figure}[h]
    \centering
    \includegraphics[width=\columnwidth]{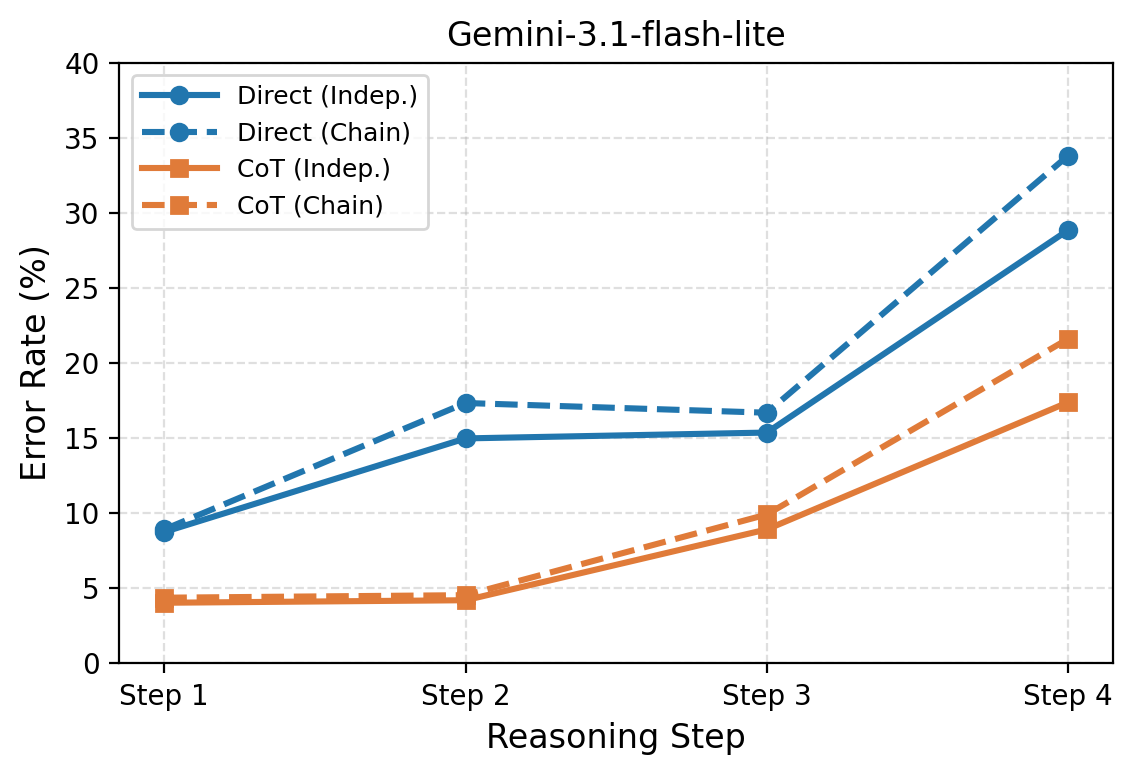}
    \caption{Step-wise error rates under independent and chain evaluation for Gemini-3.1-flash-lite.}
\end{figure}

\begin{figure}[h]
    \centering
    \includegraphics[width=\columnwidth]{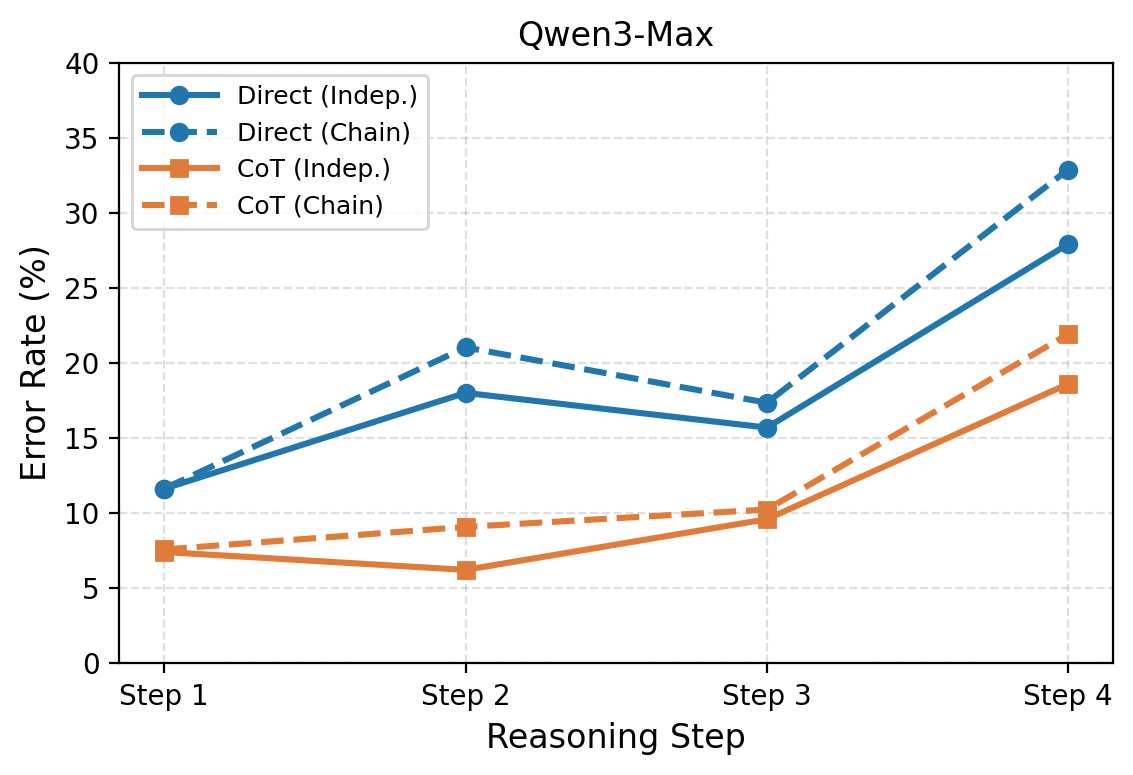}
    \caption{Step-wise error rates under independent and chain evaluation for Qwen3-Max.}
\end{figure}
\end{document}